\pdfoutput=1

\documentclass[11pt]{article}

\usepackage[]{EMNLP2022}

\usepackage{times}
\usepackage{latexsym}

\usepackage[T1]{fontenc}

\usepackage[utf8]{inputenc}

\usepackage{microtype}
\usepackage{inconsolata}
\usepackage[export]{adjustbox}
\usepackage{subcaption}
\usepackage{booktabs}
\usepackage{amssymb}
\usepackage{amsmath}
\usepackage{stfloats}
\usepackage{multirow}
\usepackage{tabularx}
\usepackage{longtable}
\usepackage{hyperref}
\usepackage{array}
\usepackage{caption}

\usepackage{url}

\usepackage{breakurl}
\usepackage[breaklinks]{hyperref}

\newcommand\shortsection[1]{\vspace{6pt}{\noindent\bf #1.}}

\usepackage{color}

\usepackage[]{trackchanges}
\addeditor{YJ}

\title{Balanced Adversarial Training: Balancing Tradeoffs between \\
Fickleness and Obstinacy in NLP Models}

\author{Hannah Chen, 
    Yangfeng Ji, David Evans \\
    Department of Computer Science\\
    University of Virginia\\
    Charlottesville, VA 22904\\
  \texttt{\{yc4dx,yangfeng,evans\}@virginia.edu} \\
 }

\begin{document}
\maketitle



\begin{abstract}

Traditional (\emph{fickle}) adversarial examples involve finding a small perturbation that does not change an input's true label but confuses the classifier into outputting a different prediction. Conversely, \emph{obstinate} adversarial examples occur when an adversary finds a small perturbation that preserves the classifier's prediction but changes the true label of an input.
Adversarial training and certified robust training have shown some effectiveness in improving the robustness of machine learnt models to fickle adversarial examples. We show that standard adversarial training methods focused on reducing vulnerability to fickle adversarial examples may make a model more vulnerable to obstinate adversarial examples, with experiments for both natural language inference and paraphrase identification tasks. To counter this phenomenon, we introduce \textit{Balanced Adversarial Training}, which incorporates contrastive learning to increase robustness against both fickle and obstinate adversarial examples.
\end{abstract}
\section{Introduction}
Interpreted broadly, an adversarial example is an input crafted intentionally to confuse a model. Most research on adversarial examples, however, focuses on a definition of an adversarial example as an input that is constructed by making minimal perturbations to a normal input that change the model's output, assuming that the small perturbations preserve the original true label~\citep{goodfellow2015explaining}. Such adversarial examples occur when a model is overly influenced by small changes in the input. 
Attackers can also target the opposite objective---to find inputs with minimal changes that change the ground truth label but for which the model retains its prior prediction~\citep{jacobsen2018excessive}.  

Various names have been used in the research literature for these two types of adversarial examples including perturbation or sensitivity-based and invariance-based examples~\citep{jacobsen2018excessive,jacobsen2019safeml}, and over-sensitive and over-stable examples~\cite{niu-bansal-2018-adversarial,Kumar2020ExplainableAE}. To avoid confusions associated with these names, we refer them as \emph{fickle adversarial examples} (the model changes its output too easily) and \emph{obstinate adversarial examples} (the model doesn't change its output even though the input has changed in a way that it should).

In NLP, synonym-based word substitution is a common method for constructing fickle adversarial examples~\citep{alzantot-etal-2018-generating, Jin_Jin_Zhou_Szolovits_2020} since synonym substitutions are assumed to not change the true label for an input.  These methods target a model's weakness of being invariant to certain types of changes which makes its predictions insufficiently responsive to small input changes. Attacks based on antonyms and negation have been proposed to create obstinate adversarial examples for dialogue models~\citep{niu-bansal-2018-adversarial}.

Adversarial training is considered as the most effective defense strategy yet found against adversarial examples~\citep{madry2018towards, cleverhans}. It aims to improve robustness by augmenting the original training set with generated adversarial examples in a way that results in decision boundaries that correctly classify inputs that otherwise would have been fickle adversarial examples. 
Adversarial training has been shown to improve robustness for NLP models~\citep{yoo-qi-2021-towards-improving}. Recent works have also studied certified robustness training which gives a stronger guarantee that the model is robust to all possible perturbations of a given input~\citep{jia-etal-2019-certified, ye-etal-2020-safer}.

\begin{figure}[tb]
\centering
\includegraphics[width=0.4\textwidth]{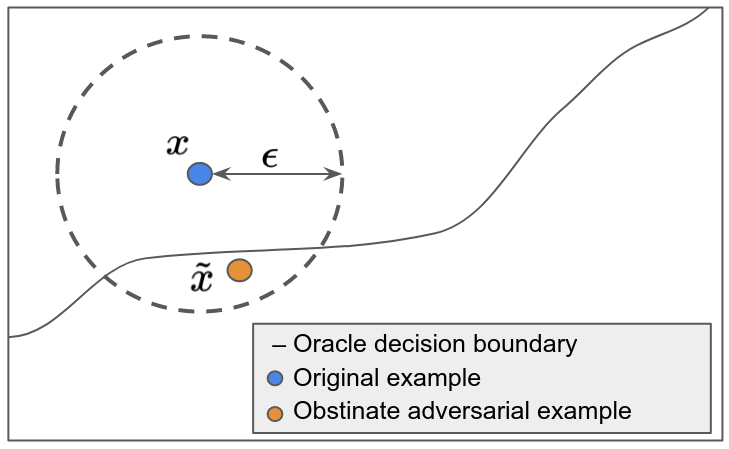}
\caption{Distance-oracle misalignment~\citep{pmlr-v119-tramer20a}. While the model is trained to be robust to $\epsilon$-bounded perturbation, it becomes too invariant to small changes in the example (obstinate example $\tilde{x}$) that lie on the other side of the oracle decision boundary.}
\label{fig:distance-misalignment}
\end{figure}

While prior work on NLP robustness focuses on fickle adversarial examples, we consider both fickle and obstinate adversarial examples. We then further examine the impact of methods designed to improve robustness to fickle adversarial examples on a model's vulnerability to obstinate adversarial examples. Recent work in the vision domain demonstrated that increasing adversarial robustness of image classification models by training with fickle adversarial examples may increase vulnerability to obstinate adversarial examples~\citep{pmlr-v119-tramer20a}. Even in cases where the model certifiably guarantees that no adversarial examples can be found within an $L_p$-bounded distance, the norm-bounded perturbation does not align with the ground truth decision boundary. This \textit{distance-oracle misalignment} makes it possible to have obstinate adversarial examples located within the same perturbation distance, as depicted in Figure~\ref{fig:distance-misalignment}. In text, fickle examples are usually generated with a cosine similarity constraint to encourage the representations of the original and the perturbed sentence to be close in the embedding space. However, this similarity measurement may not preserve the actual semantics~\citep{morris-etal-2020-reevaluating} and the model may learn poor representations during adversarial training.

\begin{figure*}
     \centering
     \includegraphics[width=1\linewidth]{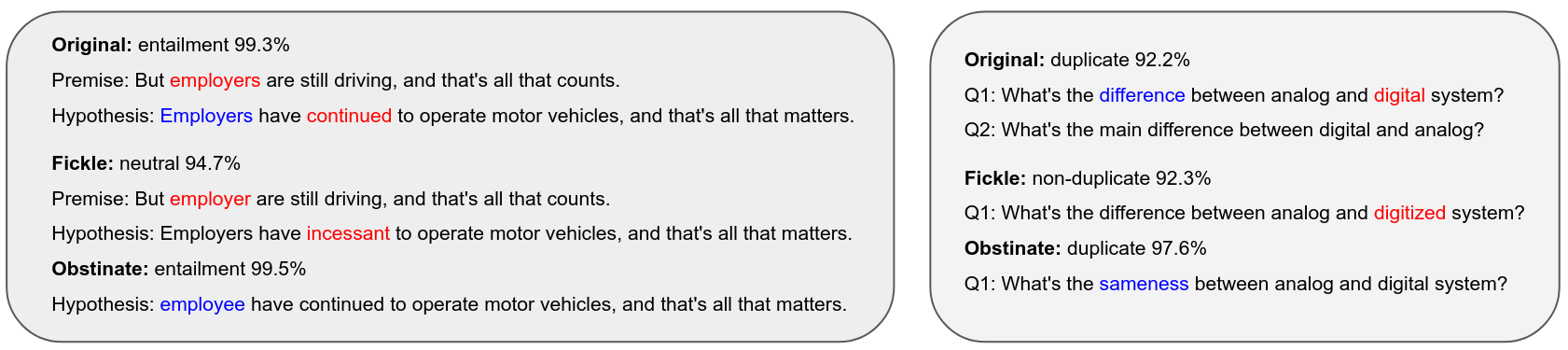}
     \caption{Fickle and obstinate adversarial examples for BERT model fine-tuned on natural language inference (left) and paraphrase identification (right) tasks. Words in red are substituted with their synonyms and words in blue are replaced by their antonyms.}
     \label{fig:adv-examples}
\end{figure*}

\shortsection{Contributions}
We study fickle and obstinate adversarial robustness in NLP models with a focus on synonym and antonym-based adversarial examples (\autoref{fig:adv-examples} shows a few examples). We evaluate both kinds of adversarial robustness on natural language inference and paraphrase identification tasks with  BERT~\citep{devlin-etal-2019-bert} and RoBERTa~\citep{liu2019roberta} models. We find that there appears to be a tradeoff between robustness to synonym-based and antonym-based attacks. We show that while certified robust training increases robustness against synonym-based adversarial examples, it increases vulnerability to antonym-based attacks (\autoref{sec:tradeoff}). We propose a modification to robust training, \textit{Balanced Adversarial Training} (BAT), which uses a contrastive learning objective to help mitigate the distance misalignment problem by learning from both fickle and obstinate examples (\autoref{sec:BAT}). We implement two versions of BAT with different contrastive learning objectives, and show the effectiveness in improving both fickleness and obstinacy robustness (\autoref{sec:BAT-results}). 


\section{Constructing Adversarial Examples}
We consider a classification task where the goal of the model $f$ is to learn to map the textual input $x$, a sequence of words, $x_1, x_2, ..., x_L$, to its ground truth label $y \in \{1,...,c\}$. We assume there is a labeling oracle $\mathcal{O}$ that corresponds to ground truth and outputs the true label of the given input. We focus on word-level perturbations where the attacker substitutes words in the original input $x$ with words from a known perturbation set (which we show how we construct it in the following sections). The goal of the attacker is to find an adversarial example $\tilde{x}$ for input $x$ such that the output of the model is different from what human would interpret, i.e. $f(\tilde{x}) \neq \mathcal{O}(\tilde{x})$. 

\subsection{Fickle Adversarial Examples}
\label{sec:oversensitive-adv}

For a given input $(x,y)$ correctly classified by model $f$ and a set of allowed perturbed sentences $\mathcal{S}_x$, an \emph{fickle adversarial example} is defined as an input $\tilde{x}_{f}$ such that:
\begin{flalign*}
&\quad \textrm{1. } \tilde{x}_{f} \in \mathcal{S}_x &\\
&\quad \textrm{2. } f(\tilde{x}_{f}) \neq f(x) &\\
&\quad \textrm{3. } \mathcal{O}(\tilde{x}_{f}) = \mathcal{O}(x) &
\end{flalign*} 

There are many different methods for finding fickle adversarial examples. The most common way is to use synonym word substitutions where the target words are replaced with similar words found in the word embedding~\citep{alzantot-etal-2018-generating, Jin_Jin_Zhou_Szolovits_2020} or use known synonyms from WordNet~\citep{ren-etal-2019-generating}. Recent work has also explored using masked language models to generate word replacements~\citep{li-etal-2020-bert-attack, garg-ramakrishnan-2020-bae, li-etal-2021-contextualized}.

We adopt the synonym word substitution method as in \citet{ye-etal-2020-safer}.
For each word $x_i$ in an input $x$, we create a synonym set $S_{x_i}$ containing the synonym words of $x_i$ including itself.  $\mathcal{S}_x$ is then constructed by a set of sentences where each word in $x$ can be replaced by a word in $S_{x_i}$. We consider the case where the attacker does not have a constraint on the number of words that can be perturbed for each input, meaning the attacker can perturb up to $L$ words which is the length of $x$. 

The underlying assumption for fickle examples to work is that the perturbed sentence $\tilde{x}_{f} \in S_{x}$ should have the same ground truth label as the original input $x$, i.e. $\mathcal{O}(\tilde{x}_{f}) = \mathcal{O}(x) = f(x)$. 
However, common practice for constructing fickle examples does not guarantee this is true. Swapping a word with its synonym may change the semantic meaning of the example since even subtle changes in words can have a big impact on meaning, and 
a word can have different meanings in different context. For instance, ``the whole 
\textit{race} of human kind'' and ``the whole \textit{competition} of human kind'' describe different things. Nonetheless, previous human evaluation has shown that synonym-based adversarial examples still retain the same semantic meaning and label as the original texts most of the time~\citep{Jin_Jin_Zhou_Szolovits_2020, li-etal-2020-bert-attack}. 

\subsection{Obstinate Adversarial Examples}

For a given input $(x,y)$ correctly classified by model $f$ and a set of allowable perturbed sentences $\mathcal{A}_x$, an \emph{obstinate adversarial example} is defined as an input $\tilde{x}_{o}$ such that:
\begin{flalign*}
&\quad \textrm{1. } \tilde{x}_{o} \in \mathcal{A}_{x} &\\
&\quad \textrm{2. } f(\tilde{x}_{o}) = f(x) &\\
&\quad \textrm{3. } \mathcal{O}(\tilde{x}_{o}) \neq \mathcal{O}(x) &
\end{flalign*}

While it is challenging to construct obstinate adversarial examples automatically for image classifiers \citep{pmlr-v119-tramer20a}, we are able to automate the process for NLP models.
We use a similar antonym word substitution strategy as proposed by \citet{niu-bansal-2018-adversarial} to construct obstinate adversarial examples. Similar to synonym word substitutions, for each word $x_i$ in an input $x$, we construct an antonym set $A_{x_i}$ that consists of the antonyms of $x_i$. Since we would like to change the semantic meaning of the input in a way that is likely to flip its label for the task, the attacker is only allowed to perturb one word with its antonym for each sentence.

The way we construct obstinate adversarial examples may not always satisfy the assumption where the ground truth label of the obstinate example would be different from the original input. The substituted word may not affect the semantic meaning of the input depending on the task. For example, in natural language inference, changing ``the weather is \textit{great}, we should go out and have fun'' to ``the weather is \textit{bad}, ...'' does not effect the entailment relationship with ``we should have some outdoor activities'' since the main argument is in the second part of the sentence. However, we find that antonym substitutions are able to change the semantic meaning of the text most of the time and we choose two tasks that are most likely to change the label under antonym-based attack.


\section{Robustness Tradeoffs}
\label{sec:tradeoff}


Normally, adversarial defense methods only target fickle adversarial examples, so there is a risk that such methods increase vulnerability to obstinate adversarial examples. 
According to the distance-oracle misalignment assumption~\citep{pmlr-v119-tramer20a} as depicted in \autoref{fig:distance-misalignment}, the distance measure for finding adversarial examples and labeling oracle $\mathcal{O}$ is misaligned if we have $\mathcal{O}(\tilde{x}_{f})=\mathcal{O}(x)=y$ and $\mathcal{O}(\tilde{x}_{o}) \neq \mathcal{O}(x)$, but $\mathit{dist}(x, \tilde{x}_{f}) > \mathit{dist}(x, \tilde{x}_{o})$. 


\subsection{Setup}
\label{sec:tradeoff-setup}
Our experiments are designed to test our hypothesis that optimizing adversarial robustness of NLP models using only fickle examples deteriorates the model's robustness on obstinate adversarial examples.  We use the SAFER certified robust training method proposed by \citet{ye-etal-2020-safer}.
The idea is to train a smoother model by randomly perturbing the sentences with words in the synonym substitution set at each training iteration. While common IBP-based certified robust training methods do not scale well onto large pre-trained language models~\citep{jia-etal-2019-certified, huang-etal-2019-achieving}, SAFER is a structure-free approach that can be applied to any kind of model architectures. In addition, it gives stronger robustness than traditional adversarial training method~\citep{yoo-qi-2021-towards-improving}.

We train BERT~\citep{devlin-etal-2019-bert} and RoBERTa~\citep{liu2019roberta} models on two different tasks with SAFER training for 15 epochs. We then test the attack success rate for both fickleness and obstinacy attacks at each training epoch. We use the same perturbation method as described in \autoref{sec:oversensitive-adv} for both the training and the attack. For each word, the synonym perturbation set is constructed by selecting the top $k$ nearest neighbors with a cosine similarity constraint of 0.8 in GLOVE embeddings~\citep{pennington-etal-2014-glove}, and the antonym perturbation set consists of antonym words found in WordNet~\citep{miller-wordnet-1995}. We follow the method of~\citet{Jin_Jin_Zhou_Szolovits_2020} for finding fickle adversarial examples by using word importance ranking and Part-of-Speech (PoS) and sentence semantic similarity constraints as the search criteria. We replace words from the ones with the highest word importance scores to the ones with the least and make sure the new substituted words have the same PoS tags as the original words. For antonym attack, we also use word importance ranking and PoS to search for word substitutions. 
For comparison, we set up baseline models with normal training on the original training sets.

\subsection{Tasks}
We choose two different tasks from the GLUE benchmark~\citep{wang-etal-2018-glue} that are good candidates for the antonym attack. Antonym-based attacks work well on these tasks since both tasks consist of sentence pairs and changing a word to an opposite meaning is likely to break the relationship between the pairs.

\shortsection{Natural Language Inference} We experiment with Multi-Genre Natural Language Inference (MNLI) dataset~\citep{williams-etal-2018-broad} which contains a premise-hypothesis pair for each example. The task is to identify the relation between the sentences in a premise-hypothesis pair and determine whether the hypothesis is true (\emph{entailment}), false (\emph{contradiction}) or undetermined (\emph{neutral}) given the premise.  We consider the case where both premise and hypothesis can be perturbed, but only one word from either premise or hypothesis can be substituted for antonym attack. We exclude examples with a \emph{neutral} label when constructing obstinate adversarial examples since antonym word substitutions may not change their label to a different class.

\shortsection{Paraphrase Identification} We use Quora Question Pairs (QQP)~\citep{qqp} which consists of questions extracted from Quora. The goal of the task is to identify duplicate questions. Each question pair is labeled as duplicate or non-duplicate. For our antonym attack strategy, we only target the duplicate class since antonym word substitutions are unlikely to flip an initially non-duplicate pair into a duplicate.

We also conducted experiments using the Wiki Talk Comments~\citep{wiki-toxic} dataset, a dataset for toxicity detection, by adding or removing toxic words for creating obstinate examples. However, we found adding toxic words can reach almost 100\% attack success rate, so there did not seem to be an interesting tradeoff to explore for available models for this task, and we do not include it in our results.

\begin{figure*}[ht]
     \centering
     \begin{subfigure}[b]{0.43\linewidth}
         \centering
         \includegraphics[width=1.25\textwidth]{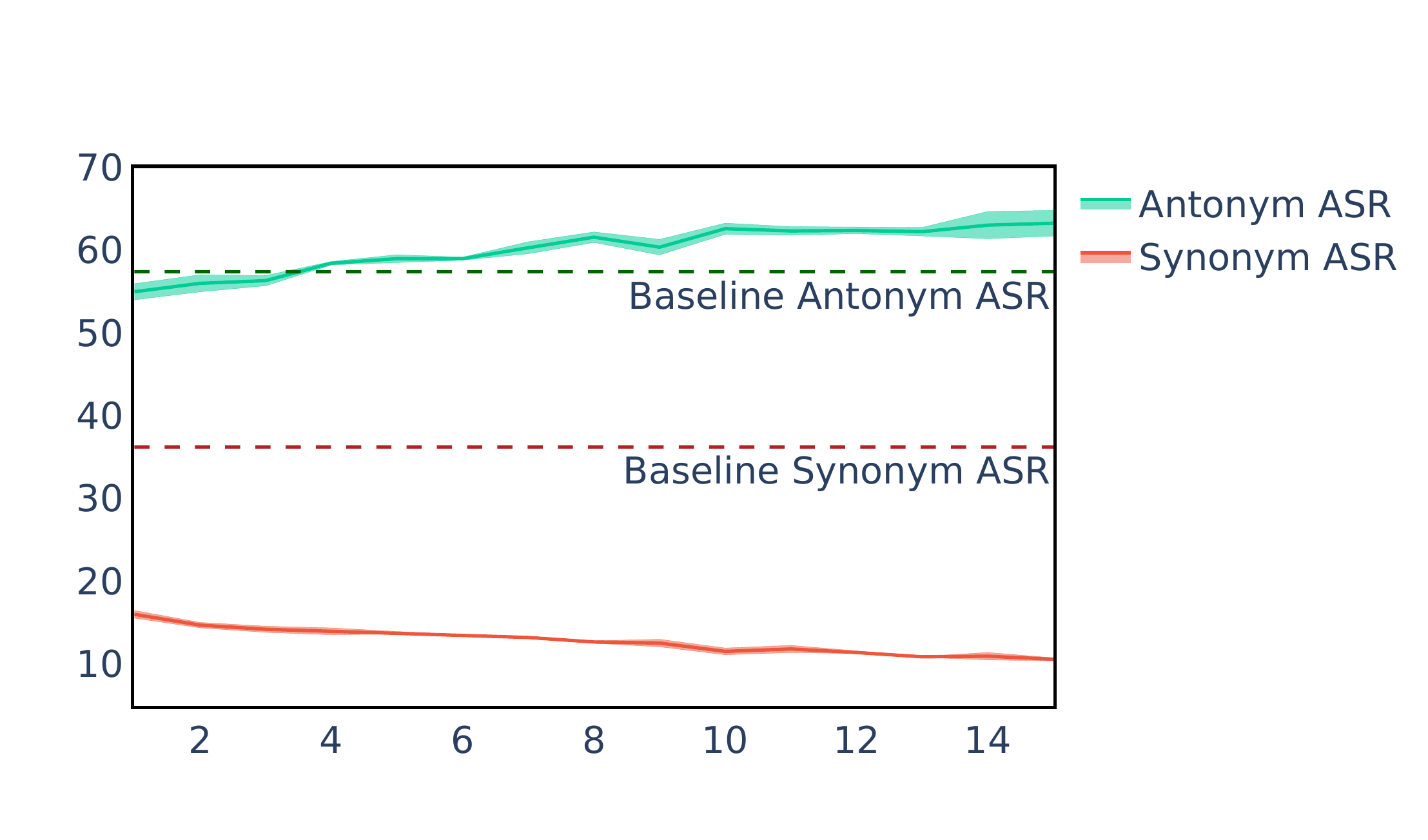}
         \caption{MNLI (BERT)}
         \label{fig:bert-mnli-tradeoff}
     \end{subfigure}
     \hfill
     \begin{subfigure}[b]{0.43\linewidth}
         \centering
         \includegraphics[width=1.05\textwidth]{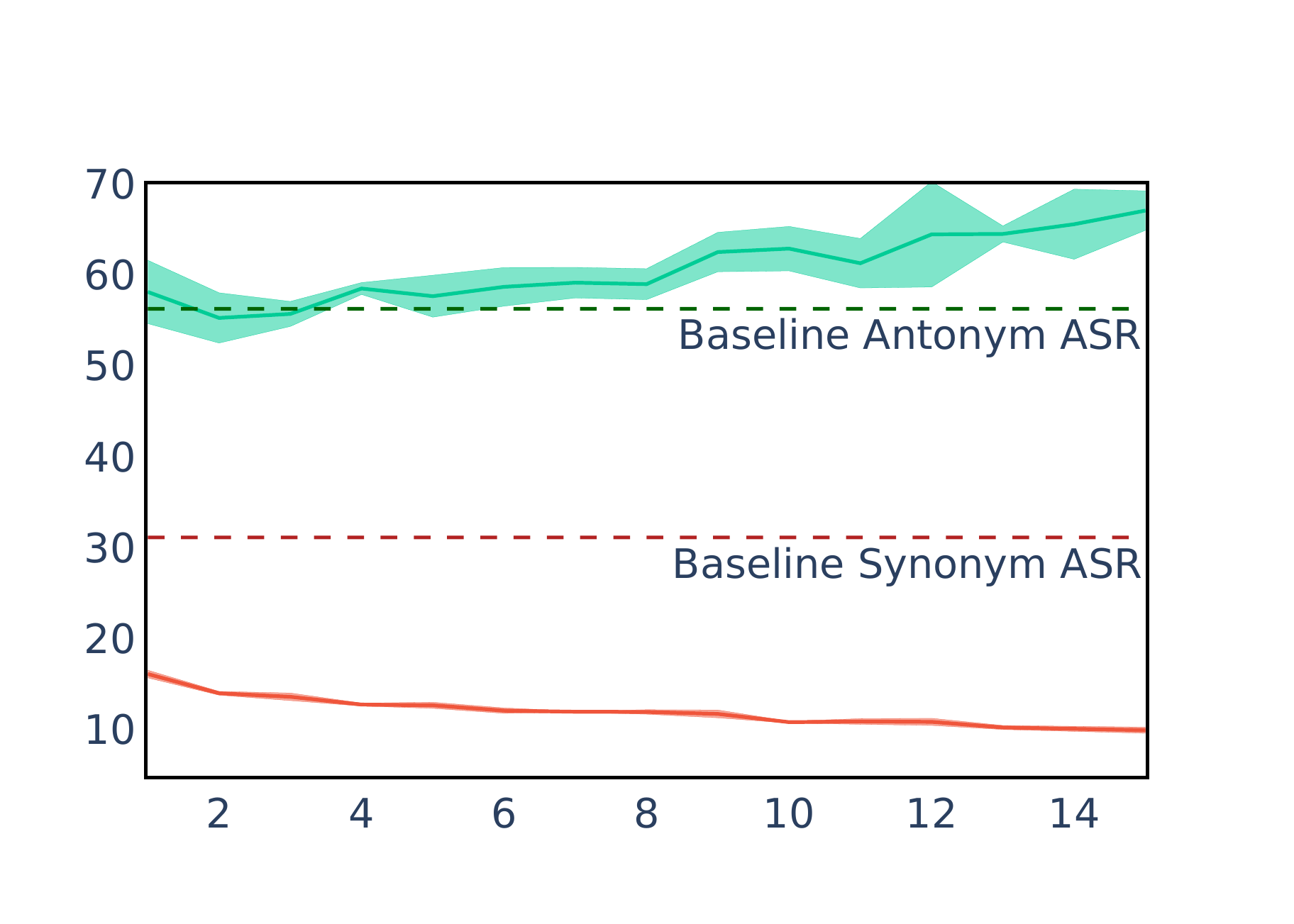}
         \caption{MNLI (RoBERTa)}
         \label{fig:roberta-mnli-tradeoff}
     \end{subfigure}
     \caption{Fickleness and obstinacy tradeoff where fickleness attack success rate increases as obstinacy attack success rate decreases. The figure shows the results on MNLI matched validation set with average and standard deviation across three different runs. Dash lines show the synonym/antonym attack success rate on baseline model with normal training.}
     \label{fig:tradeoff-mnli}
\end{figure*}

\begin{figure*}
    \centering
    \begin{subfigure}[b]{0.42\linewidth}
         \centering
         \includegraphics[width=1.2\textwidth]{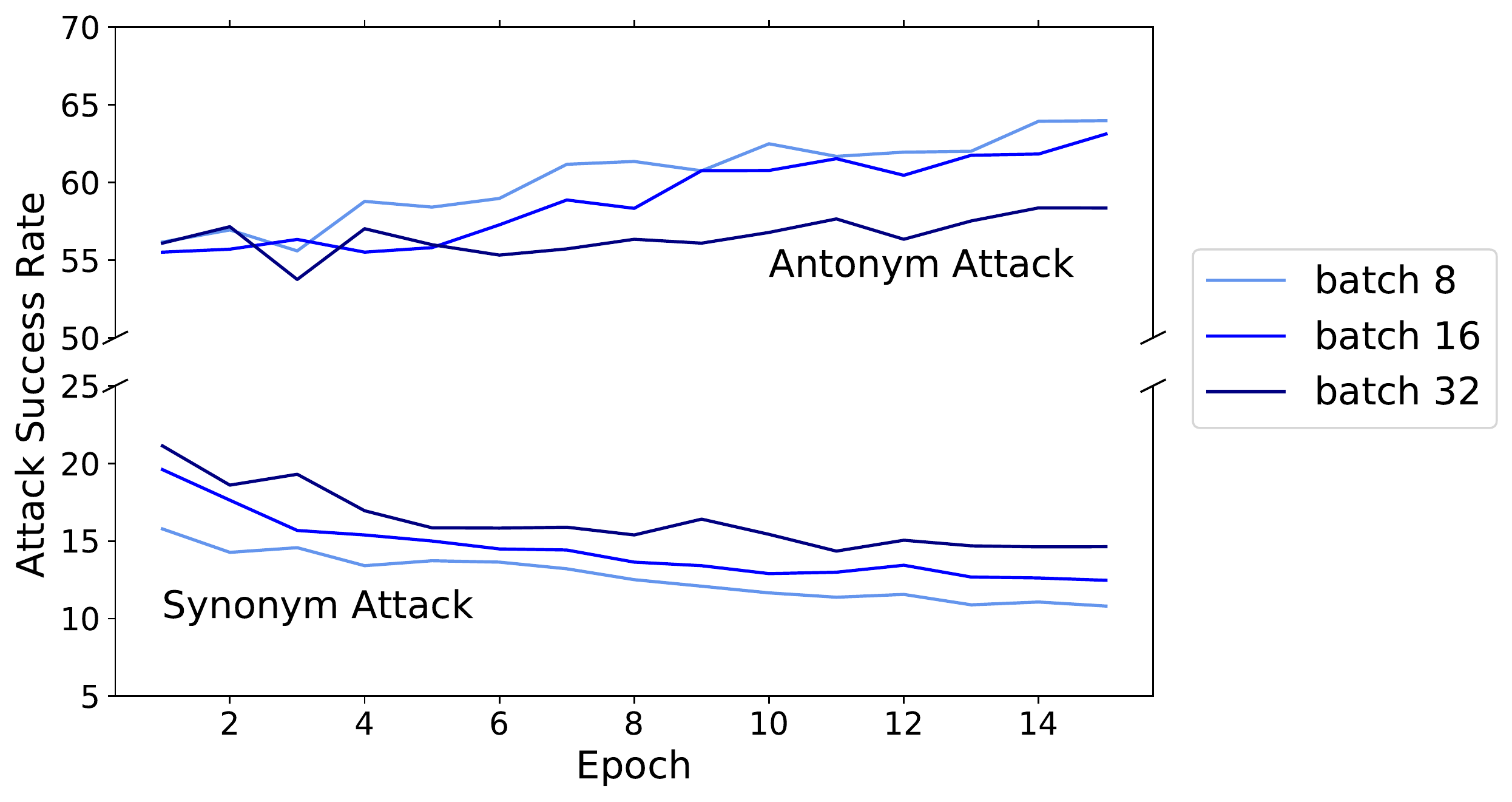}
         \caption{MNLI (BERT)}
         \label{fig:bert-mnli-batch-size-asr}
     \end{subfigure}
     \hfill
     \begin{subfigure}[b]{0.42\linewidth}
         \centering
         \includegraphics[width=0.95\textwidth]{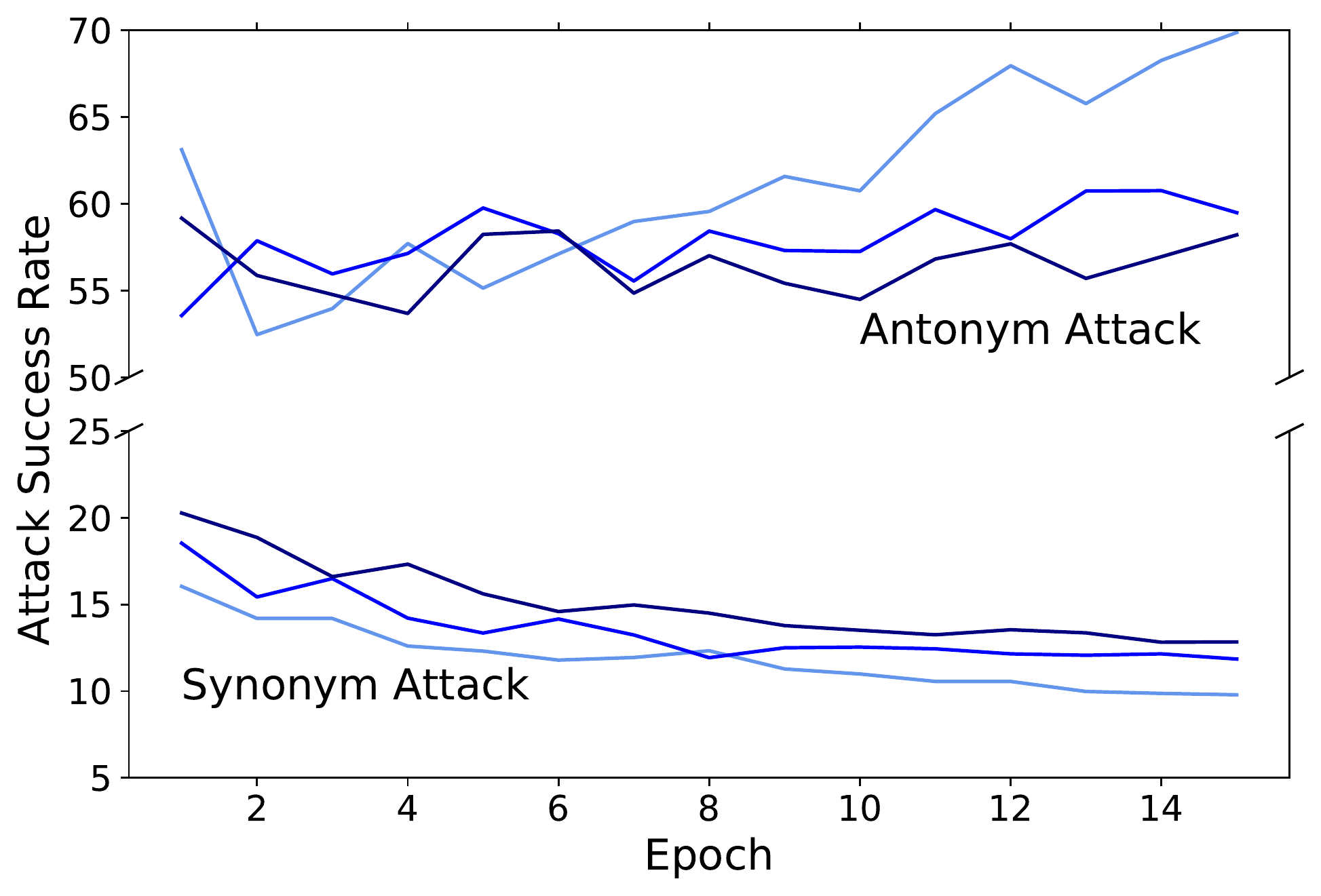}
         \caption{MNLI (RoBERTa)}
         \label{fig:roberta-mnli-batch-size-asr}
     \end{subfigure}
    \caption{The synonym and antonym attack success rate at each SAFER training epoch with varying batch size. When the model is trained with smaller batch size, the synonym attack success rate is lower and the antonym success rate is higher.}
    \label{fig:batch-size-asr}
\end{figure*}

\subsection{Results}
\label{sec:tradeoff-results}

We visualize the attack success rates for fickleness (synonym attack) and obstinacy (antonym attack) attacks in \autoref{fig:tradeoff-mnli}.
The results are consistent with our hypothesis that optimizing adversarial robustness of NLP models using only fickle examples can result in models that are more vulnerable to obstinacy attacks. Robustness training for the BERT model on MNLI improves fickleness robustness, reducing the synonym attack success rate from 36\% to 11\% (a 69\% decrease) after training for 15 epochs (\autoref{fig:bert-mnli-tradeoff}), but antonym attack success rate increases from 56\% to 63\% (a 13\% increase).
The antonym attack success rate increases even more for the RoBERTa model (\autoref{fig:roberta-mnli-tradeoff}), increasing from 56\% to 67\% (a 20\% increase) while the synonym attack success rate decreases from 31.2\% to 10\% (a 68\% decrease). The RoBERTa model is pre-trained to be more robust than the BERT model with dynamic masking, which perhaps explains the difference. We observe a robustness tradeoff
for QQP dataset as well (see Appendix~\ref{app:qqp-tradeoff}). In addition, the fickle adversarial training does not sacrifice the performance on the original examples and increases consistently throughout training (see \autoref{fig:batch-size-eval-acc} in the appendix).



\shortsection{Impact of Batch Size}
We experiment with different batch sizes for fickle-based robust training. \autoref{fig:batch-size-asr} shows the results on MNLI dataset. When the model is trained with a smaller batch size, the synonym attack success rate becomes lower, but the antonym success rate gets higher. This means that the model may overfit on the fickle examples due to smaller training batch size, exacerbating the impact of the unbalanced adversarial training. Similar observation is found on the QQP task (see \autoref{fig:batch-size-asr-qqp} in the appendix). We found similar evidence on the evaluation accuracy on the original validation set (see \autoref{fig:batch-size-eval-acc} in the appendix). While models with smaller batch sizes converge faster, they lead to lower performance and poorer generalization. 
In Appendix~\ref{app:bat-varying-batch-size}, we show that our proposed method is not affected by the training batch size.


\section{Balanced Adversarial Training}
\label{sec:BAT}
In previous section, we argued that the tradeoff between fickleness and obstinacy can be attributed to distance-oracle misalignment. This section proposes and evaluates a modification to adversarial training that balances both kinds of adversarial examples.

\begin{table*}[tb]
    \centering
    \begin{tabular}{c c c c c}
    \toprule
    Model & Method & Eval Acc (\%) & Antonym ASR (\%) & Synonym ASR (\%) \\
    \midrule
         & Normal Training & 84.01 $\pm0.32$ & 57.48 $\pm0.14$ & 36.63 $\pm0.34$ \\
         & A2T & 84.44 & 56.51 & 21.67 \\
        BERT & SAFER & 83.81 $\pm0.40$ & 63.33 $\pm1.55$ & 10.61 $\pm0.17$ \\
         & BAT-Pairwise & 84.03 $\pm0.21$ & 35.47 $\pm1.51$ & 26.70 $\pm0.78$ \\
         & BAT-Triplet & 84.66 $\pm0.12$ & 34.25 $\pm2.19$ & 25.70 $\pm0.13$  \\
    \midrule
         & Normal Training & 87.74 $\pm0.09$ & 55.86 $\pm0.47$ & 30.86 $\pm0.50$ \\
         & A2T & 86.98 & 56.84 & 19.78 \\
        RoBERTa & SAFER & 86.18 $\pm0.18$ & 67.15 $\pm2.17$ & 10.00 $\pm0.34$ \\
         & BAT-Pairwise & 87.23 $\pm0.38$ & 38.68 $\pm1.61$ & 27.29 $\pm0.20$ \\
         & BAT-Triplet & 87.44 $\pm0.12$ & 33.02 $\pm0.81$ & 27.54 $\pm0.45$ \\
    \bottomrule
    \end{tabular}
    \caption{Balanced Adversarial Training evaluation results on MNLI matched validation set. Results shown with standard deviations are average across three different runs.}
    \label{tab:balanced-adv-train-mnli}
\end{table*}

\subsection{Approach}

The most intuitive way to make the semantic distance in the representation space align better with human perception is to move the fickle example closer to the original input and push the original input apart from the obstinate example in the representation space.
%
This goal matches the objective of contrastive learning, a type of self-supervised learning that learns representations with positive (similar) examples close together and negative (dissimlar) examples far apart~\citep{hadsell2006dimensionality, schroff2015facenet}. Positive examples are usually generated with data augmentation such as spatial transformation, and negative examples are sampled from other examples~\citep{chen2020simple}. 

We adapt contrastive learning to balance adversarial training by treating fickle adversarial examples as positive examples and obstinate adversarial examples as negative examples. The idea is to minimize the distance between the positive pairs and maximize the distance between the negative pairs. We construct positive pairs by pairing the original input with a corresponding fickle example, and negative pairs as the original input paired with an obstinate example. We generate fickle examples by applying synonym transformations, and obstinate examples by applying antonym transformations.

We combine normal training with a contrastive learning objective and experiment with two different approaches for contrastive loss: pairwise and triplet loss. While recent contrastive learning incorporates multiple positive and negative examples for each input, we use these two methods as they consider the simplest case where only a positive and a negative example is needed for each input.
Similarly to SAFER certified robust training, we use an augmented approach without querying the model to check if the attack succeeds. We choose this approach over traditional adversarial training since it is computationally less expensive. 

Given an input $(x,y)$,  we generate an example $\tilde{x}_{o}$ by applying synonym perturbations and an example $\tilde{x}_{u}$ by applying antonym perturbations. Let $d(x_1, x_2)$ denote the distance measure between $x_1$ and $x_2$ in the representation space. 
%
%

\shortsection{BAT-Pairwise}
For the pairwise approach, we independently optimize the distance for the fickle pair $(x, \tilde{x}_{f})$ and the obstinate pair $(x, \tilde{x}_{o})$:
\begin{align*}
    \mathcal{L}^{\mathit{BAT}_{\mathit{pair}}} &= \mathcal{L}_{\mathit{ML}} + \mathcal{L}_{\mathit{pair}} \\
    \mathcal{L}_{\mathit{ML}} &= \log f(y \; | \; x) \\
    \mathcal{L}_{\mathit{pair}} &= \alpha d(x,\tilde{x}_{f}) + \beta \max(0, m-d(x,\tilde{x}_{o}))
\end{align*}
\noindent
The hyperparameters $\alpha$ and $\beta$ control the weighting of the fickle and obstinate pairs, and $m$ is the margin. The $\mathcal{L}_{pair}$ loss term is designed to minimize the distance to the fickle adversarial example and maximize the distance to the obstinate adversarial example. The margin $m$ penalizes the model when the obstinate example within $m$ distance of the original input ($d(x,\tilde{x}_{o}) < m$). We use cosine similarity for distance measure (ranges from 0 to 1), and set the margin as 1 as we find it gives the best performance (see Appendix~\ref{app:BAT-margin-search}).
For the case where we are unable to find a valid fickle and obstinate adversarial example, we set the corresponding term, either $d(x,\tilde{x}_{f})$ or $m - d(x,\tilde{x}_{o})$, to 0.


\shortsection{BAT-Triplet}
For the triplet approach, the original input $x$ acts as an anchor and a triplet, $(x,\tilde{x}_{f},\tilde{x}_{o})$, is considered instead of pairs. The triplet loss aims to make the distance between the obstinate pair larger than the distance between the fickle pair, with at least a margin $m$: $d(x,\tilde{x}_{o}) > d(x,\tilde{x}_{f}) + m$. The training loss can be formalized as:
\begin{align*}
    \mathcal{L}^{\mathit{BAT}_{\mathit{triplet}}} &= \mathcal{L}_{ML} + \lambda \mathcal{L}_{\mathit{triplet}} \\
    \mathcal{L}_{\mathit{triplet}} 
                        &= \max(0, d(x,\tilde{x}_{f}) + (m - d(x,\tilde{x}_{o})))
\end{align*}
\noindent where the hyperparameter $\lambda$ controls the weight of the contrastive loss term. Similarly to the pairwise loss, if no fickle and obstinate example is available, we mask out $d(x,\tilde{x}_{f})$ or $m - d(x,\tilde{x}_{o})$ in $\mathcal{L}_{\mathit{triplet}}$. 

We show the training details and how we find the best hyperparameters in Appendix~\ref{app:bat-hyperparameter-search}. 

\begin{table*}[tb]
    \centering
    \begin{adjustbox}{width=\textwidth}
    \begin{tabular}{c c c c c c}
    \toprule
    Model & Method & Eval Acc (\%) & F1 & Antonym ASR (\%) & Synonym ASR (\%) \\
    \midrule
         & Normal Training & 90.78 $\pm0.12$ & 87.67 $\pm0.12$ & 44.94 $\pm1.24$ & 21.16 $\pm0.36$ \\
         & A2T & 90.70 & 87.52 & 36.27 & 14.71 \\
        BERT  & SAFER & 90.85 $\pm0.19$ & 87.65 $\pm0.23$ & 50.10 $\pm0.44$ & 4.83 $\pm0.15$ \\
          & BAT-Pairwise & 90.31 $\pm0.03$ & 86.97 $\pm0.09$ & 22.46 $\pm1.08$ & 16.00 $\pm0.73$ \\
         & BAT-Triplet & 90.82 $\pm0.11$ & 87.79 $\pm0.04$ & 16.24 $\pm1.41$ & 15.73 $\pm0.09$ \\
    \midrule
         & Normal Training & 91.18 $\pm0.09$ & 88.29 $\pm0.12$ & 40.52 $\pm0.71$ & 18.9 $\pm0.12$ \\
         & A2T & 91.14 & 88.04 & 42.42 & 13.04 \\
        RoBERTa  & SAFER & 91.26 $\pm0.10$ & 88.27 $\pm0.15$ & 46.07 $\pm1.44$ & 4.87 $\pm0.45$ \\
         & BAT-Pairwise & 90.19 $\pm0.29$ & 86.94 $\pm0.40$ & 15.29 $\pm2.62$ & 16.58 $\pm0.44$ \\
         & BAT-Triplet & 91.04 & 88.21 & 13.02 & 16.89 \\
    \bottomrule
    \end{tabular}
    \end{adjustbox}
    \caption{Balanced Adversarial Training evaluation results on QQP validation set.}
    \label{tab:balanced-adv-train-qqp}
\end{table*}

\subsection{Results}
\label{sec:BAT-results}


\autoref{tab:balanced-adv-train-mnli} shows BAT training results on the MNLI validation sets. We use normal training as the non-robust baseline, and include two robust baselines: certified robust training (SAFER), and traditional adversarial training (A2T)~\citep{yoo-qi-2021-towards-improving}. 
Balanced Adversarial Training increases the model's adversarial robustness against both antonym and synonym attacks, while preserving its performance on the original validation set. 

While both of the robust baselines (SAFER and A2T), which only consider fickle adversarial examples, perform best when evaluated solely based on fickleness robustness, they are more vulnerable to obstinate adversarial examples. We found that BAT-Triplet performs better than BAT-Pairwise in terms of improving robustness against antonym attacks. 
With BAT-Triplet, the antonym attack success rate on BERT decreases from 57\% to 34\% (a 40\% decrease) comparing to normal training, and the synonym atack success rate decreases from 36\% to 26\% (a 28\% decrease).


Results for the QQP dataset are shown in \autoref{tab:balanced-adv-train-qqp}. While the antonym attack success rates drop more than half (around 67\% decrease) after BAT training, the synonym attack success rate has a 24\% decrease on BERT and only 10\% on RoBERTa, as the synonym attack success rate is already low on the model with normal training.

\subsection{Representation Analysis}
\begin{figure*}
     \centering
     \begin{subfigure}[b]{0.45\linewidth}
         \centering
         \includegraphics[width=1.2\textwidth]{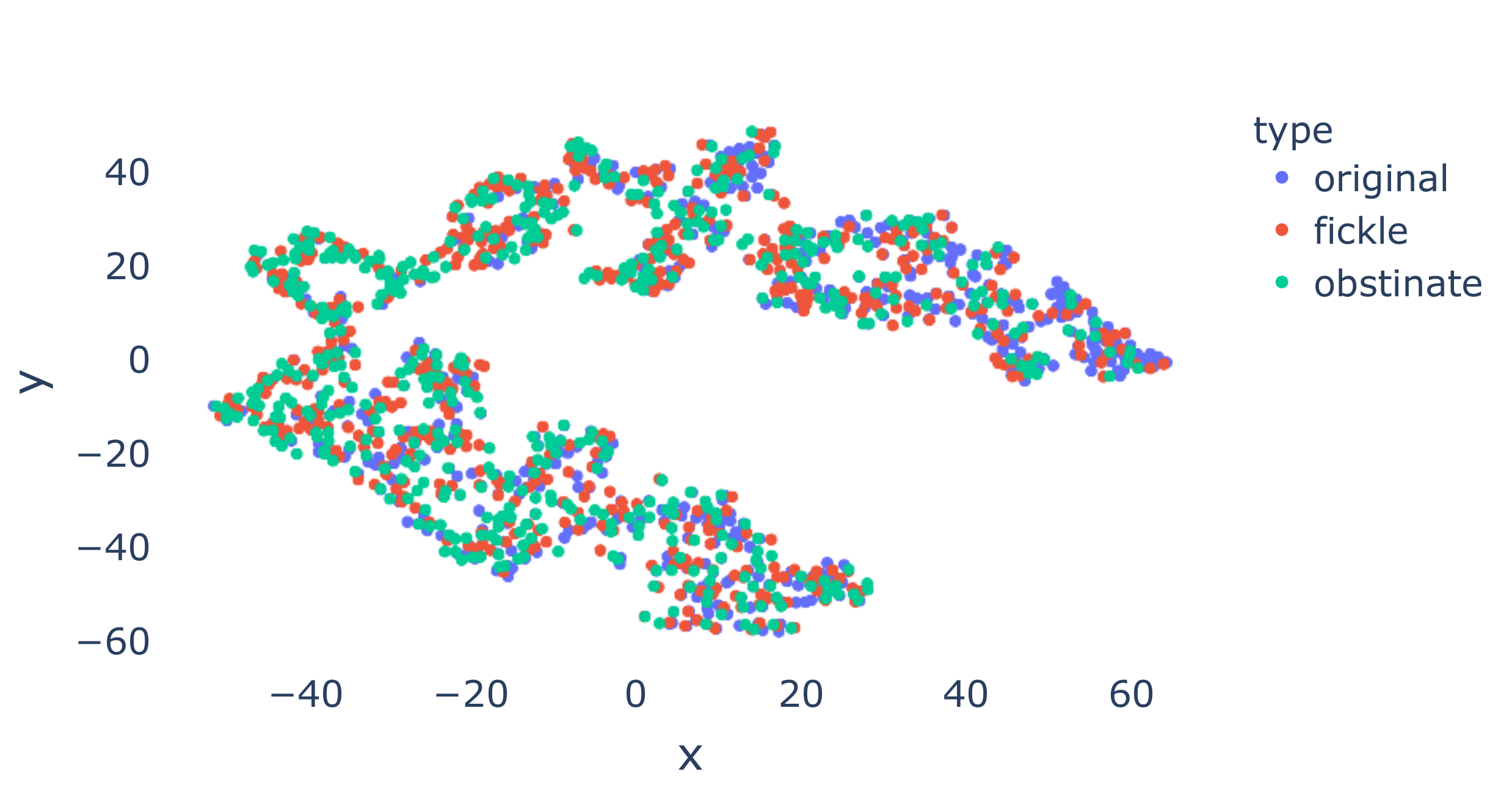}
         \caption{Normal training}
         \label{fig:embedding-baseline}
     \end{subfigure}
     \hfill
     \begin{subfigure}[b]{0.45\linewidth}
         \centering
         \includegraphics[width=1.1\textwidth]{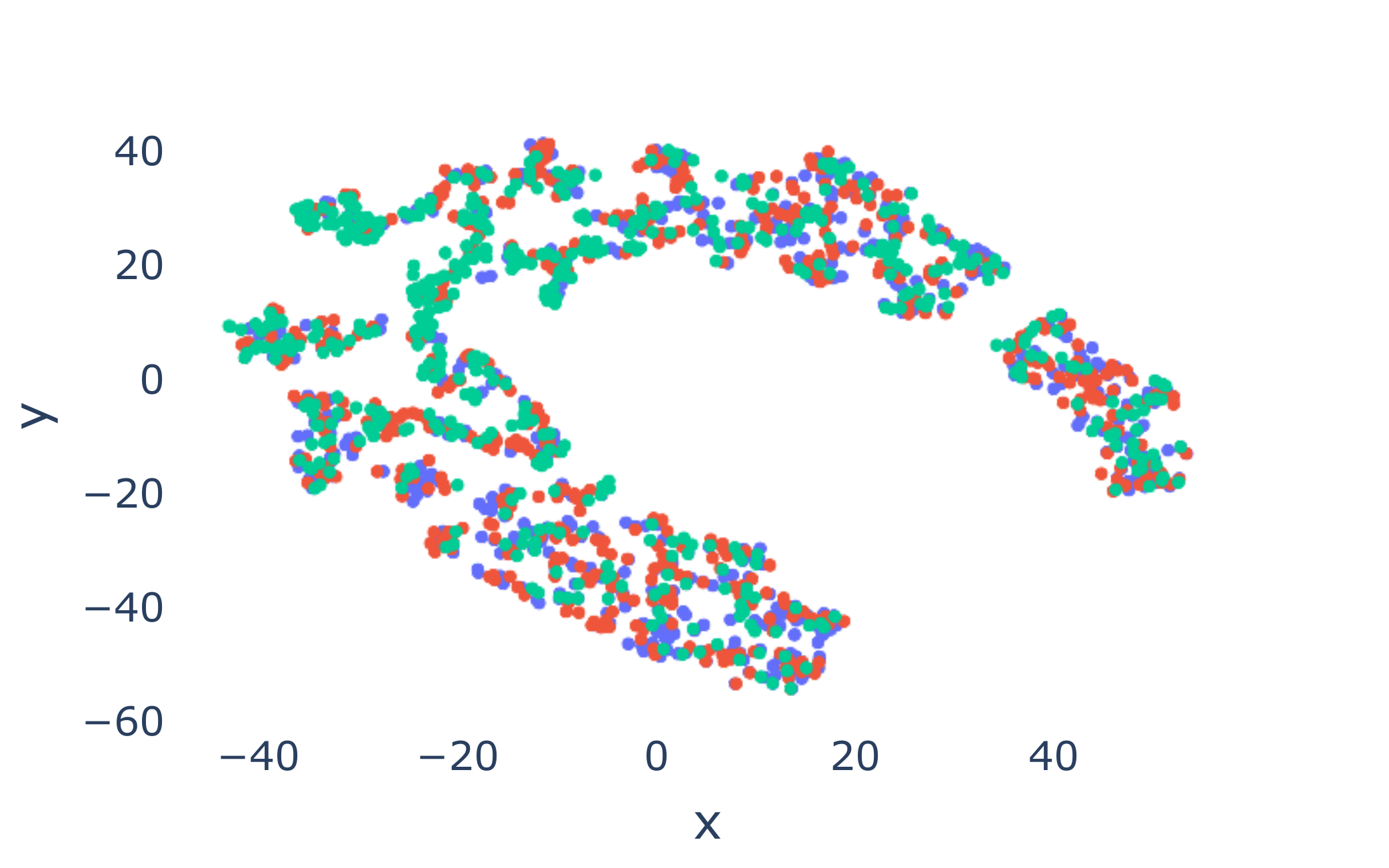}
         \caption{SAFER}
         \label{fig:embedding-safer}
     \end{subfigure}
     \hfill
     \begin{subfigure}[b]{0.45\linewidth}
         \centering
         \includegraphics[width=1.1\textwidth]{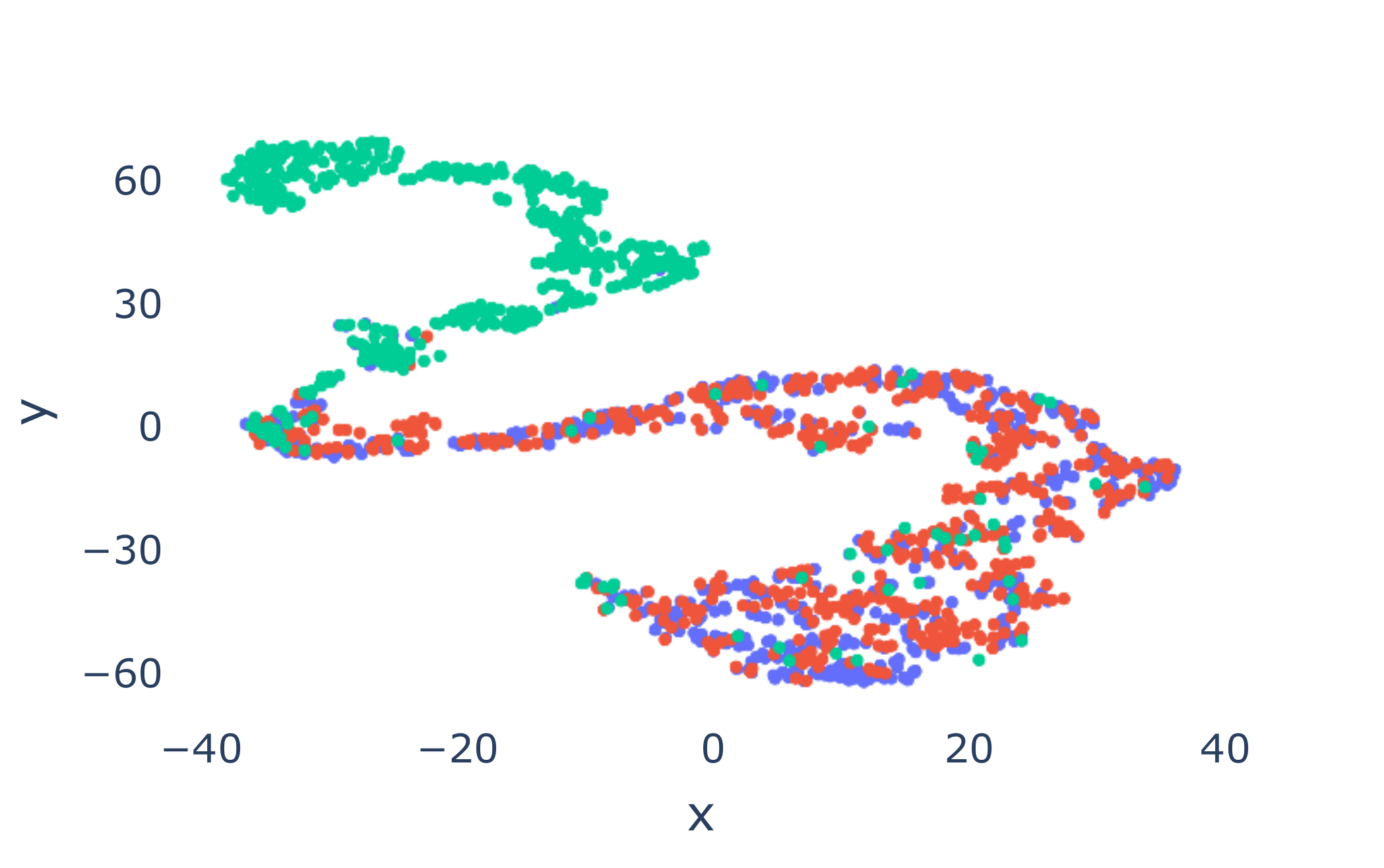}
         \caption{BAT-Pairwise}
         \label{fig:embedding-pairwise}
     \end{subfigure}
     \hfill
     \begin{subfigure}[b]{0.45\linewidth}
         \centering
         \includegraphics[width=1.1\textwidth]{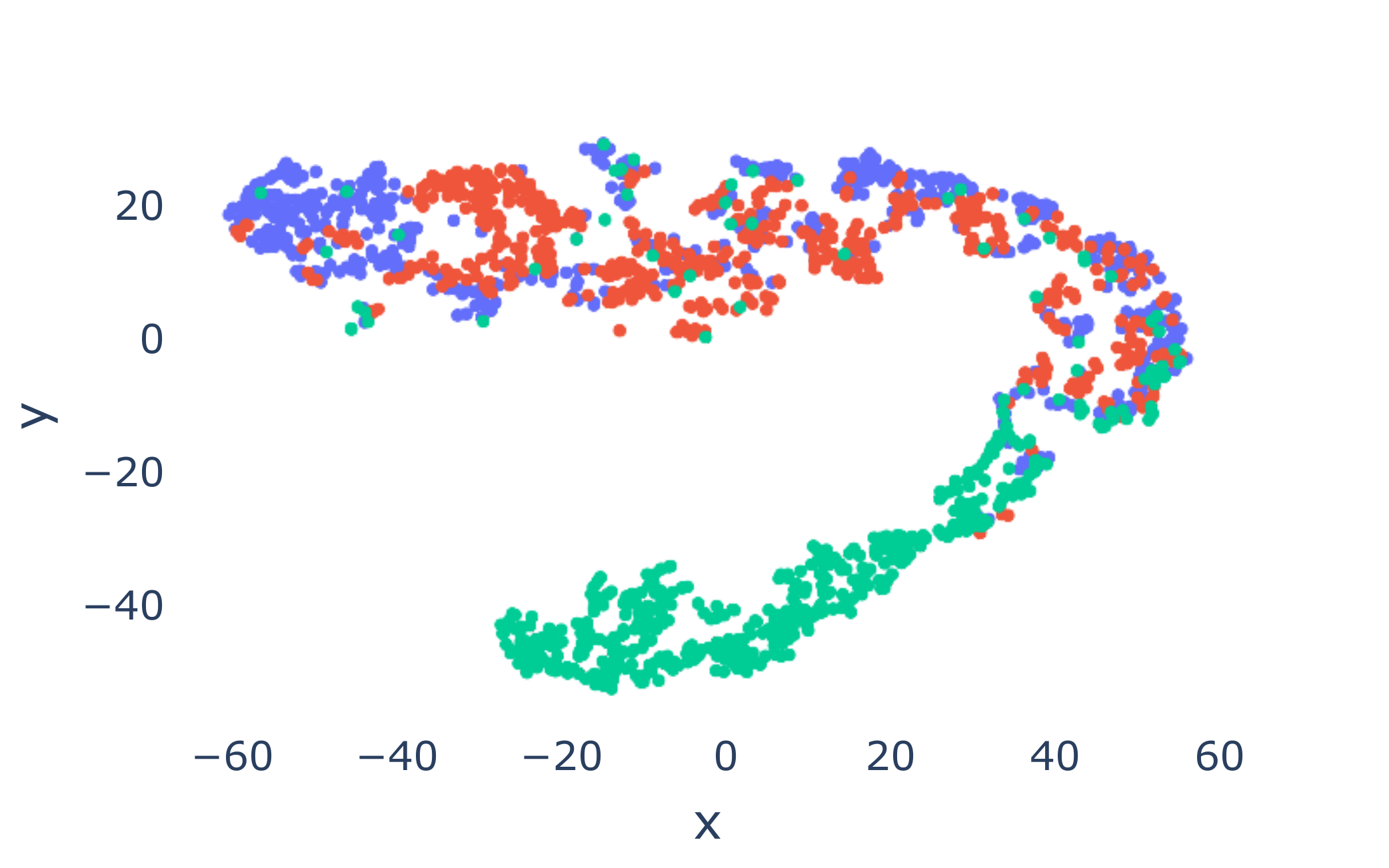}
         \caption{BAT-Triplet}
         \label{fig:embedding-triplet}
     \end{subfigure}
     \caption{2D projection of model representation for RoBERTa MNLI models trained with normal training, certified robust training with fickle adversarial examples (SAFER), BAT-Pairwise, and BAT-Triplet.}
     \label{fig:embedding-plots}
\end{figure*}

We compare the learned representations of models trained with BAT to normal training and SAFER.  We sample 500 examples from MNLI dataset (excluding the neutral class) and apply synonym and antonym perturbations for each input. We then project the model representations before the last classification layer to 2 dimensional space with t-SNE~\citep{tsne-vandermaaten08a} and visualize the results in \autoref{fig:embedding-plots}. 

When training with normal training or SAFER, we can see that both fickle and obstinate adversarial examples are fairly close to the original examples. However, with BAT-Pairwise or BAT-Triplet, obstinate examples are pushed further away from both original and fickle examples. This matches with BAT's training goal where the distance between obstinate and original examples is maximized and the distance between fickle and original examples is minimized.  This also shows how BAT is able to fix the distance-oracle misalignment, making the semantic distance in the representation space aligns better with human perception, and further improve robustness against both types of adversarial examples.

\section{Related Work}

Compared to fickle adversarial examples, obstinacy has been less studied in NLP as well as other domains. \citet{feng-etal-2018-pathologies} delete words iteratively from the input to create examples that appear rubbish to human but retain the model's prediction with high confidence. \citet{welbl-etal-2020-undersensitivity} use Part-of-Speech and Name Entity based perturbations against reading comprehension models. \citet{niu-bansal-2018-adversarial} study both types of attack strategies for dialogue models. They create obstinate adversarial examples by substituting words with antonyms or adding negation words to the input. 

Our work is the first to study tradeoffs between fickle and obstinate adversarial examples in NLP, but a few previous works have considered these tradeoffs in the vision domain.
\citet{jacobsen2018excessive} show that adversary can not only target the model's excessive sensitivity but its excessive invariance to small changes in the input. They propose an alternative training objective based on information theory to make the model less invariant to semantically meaningful changes. \citet{pmlr-v119-tramer20a} study the tradeoff between the two types of adversarial examples for image classifiers. They show that data augmentation can help increase robustness against obstinacy attacks, but is not sufficient to impede both types of attacks. Our work differs in that we propose a new adversarial training method that improves model robustness against both types of adversarial examples. In addition, unlike images where human inspection is usually required to check whether the perturbed pixels would change the true label of the image, we are able to automate the process of generating obstinate examples for text.

Recent work introduce contrastive learning for image classifiers in the adversarial learning setting where an fickle adversarial augmentation is used to generate positive examples and negative examples are sampled from other images. \citet{Kim2020adversarial} generate diverse positive examples by launching instance-wise attack on augmented images and show that it improves model's fickleness robustness. \citet{ho2020contrastive} create challenging positive pairs by using the gradients of the contrastive loss to generate fickle adversarial examples and they show that it improves model performance.


\section{Conclusion}
We demonstrate the tradeoff between vulnerability to synonym-based (fickle) and antonym-based (obstinate) adversarial examples for NLP models and show that increasing robustness against synonym based attacks also increases vulnerability to antonym-based attacks. To manage this tension, we introduce a new adversarial training method, BAT, which targets the distance-oracle misalignment problem and can help balance the fickleness and obstinacy in adversarial training. 

\section*{Limitations}
We showed robustness tradeoffs exist between synonym and antonym-based adversarial examples. Since there are numerous ways to construct adversarial examples for NLP models, further investigation is needed to show if this holds true for any kind of fickleness and obstinacy attacks for NLP models and we will leave it for future work. In order to launch antonym attack automatically, we are also limited to sentence pair tasks that are more prone to changing the ground truth label when replacing a word with its antonym. In addition, BAT sacrifices the performance on the synonym-based attack success rate for robustness to antonym-based attack when comparing to fickle adversarial training methods. We show that there is a tradeoff between robustness against synonym and antonym based attacks and our goal is to achieve a better tradeoff between them.

\section*{Availability}
Code for reproducing our experiments is available at: \url{https://github.com/hannahxchen/balanced-adversarial-training}.

\section*{Acknowledgements}

This work was partially supported by a grant from the National Science Foundation (\#1804603).

\bibliography{anthology,custom}

\begin{thebibliography}{35}
\expandafter\ifx\csname natexlab\endcsname\relax\def\natexlab#1{#1}\fi

\bibitem[{Alzantot et~al.(2018)Alzantot, Sharma, Elgohary, Ho, Srivastava, and
  Chang}]{alzantot-etal-2018-generating}
Moustafa Alzantot, Yash Sharma, Ahmed Elgohary, Bo-Jhang Ho, Mani Srivastava,
  and Kai-Wei Chang. 2018.
\newblock \href {https://doi.org/10.18653/v1/D18-1316} {Generating natural
  language adversarial examples}.
\newblock In \emph{Proceedings of the 2018 Conference on Empirical Methods in
  Natural Language Processing}.

\bibitem[{Chen et~al.(2020)Chen, Kornblith, Norouzi, and
  Hinton}]{chen2020simple}
Ting Chen, Simon Kornblith, Mohammad Norouzi, and Geoffrey~E. Hinton. 2020.
\newblock \href {https://arxiv.org/abs/2002.05709} {A simple framework for
  contrastive learning of visual representations}.
\newblock \emph{CoRR}, abs/2002.05709.

\bibitem[{Devlin et~al.(2019)Devlin, Chang, Lee, and
  Toutanova}]{devlin-etal-2019-bert}
Jacob Devlin, Ming-Wei Chang, Kenton Lee, and Kristina Toutanova. 2019.
\newblock \href {https://doi.org/10.18653/v1/N19-1423} {{BERT}: Pre-training of
  deep bidirectional transformers for language understanding}.
\newblock In \emph{Proceedings of the 2019 Conference of the North {A}merican
  Chapter of the Association for Computational Linguistics: Human Language
  Technologies, Volume 1 (Long and Short Papers)}.

\bibitem[{Feng et~al.(2018)Feng, Wallace, Grissom~II, Iyyer, Rodriguez, and
  Boyd-Graber}]{feng-etal-2018-pathologies}
Shi Feng, Eric Wallace, Alvin Grissom~II, Mohit Iyyer, Pedro Rodriguez, and
  Jordan Boyd-Graber. 2018.
\newblock \href {https://doi.org/10.18653/v1/D18-1407} {Pathologies of neural
  models make interpretations difficult}.
\newblock In \emph{Proceedings of the 2018 Conference on Empirical Methods in
  Natural Language Processing}.

\bibitem[{Garg and Ramakrishnan(2020)}]{garg-ramakrishnan-2020-bae}
Siddhant Garg and Goutham Ramakrishnan. 2020.
\newblock \href {https://doi.org/10.18653/v1/2020.emnlp-main.498} {{BAE}:
  {BERT}-based adversarial examples for text classification}.
\newblock In \emph{Proceedings of the 2020 Conference on Empirical Methods in
  Natural Language Processing (EMNLP)}.

\bibitem[{Goodfellow et~al.(2016)Goodfellow, Papernot, and
  McDaniel}]{cleverhans}
Ian~J. Goodfellow, Nicolas Papernot, and Patrick~D. McDaniel. 2016.
\newblock \href {http://arxiv.org/abs/1610.00768} {cleverhans v0.1: an
  adversarial machine learning library}.
\newblock \emph{CoRR}, abs/1610.00768.

\bibitem[{Goodfellow et~al.(2015)Goodfellow, Shlens, and
  Szegedy}]{goodfellow2015explaining}
Ian~J. Goodfellow, Jonathon Shlens, and Christian Szegedy. 2015.
\newblock \href {https://arxiv.org/abs/1412.6572} {Explaining and harnessing
  adversarial examples}.
\newblock \emph{CoRR}, abs/1412.6572.

\bibitem[{Hadsell et~al.(2006)Hadsell, Chopra, and
  LeCun}]{hadsell2006dimensionality}
R.~Hadsell, S.~Chopra, and Y.~LeCun. 2006.
\newblock \href {https://doi.org/10.1109/CVPR.2006.100} {Dimensionality
  reduction by learning an invariant mapping}.
\newblock In \emph{2006 IEEE Computer Society Conference on Computer Vision and
  Pattern Recognition (CVPR'06)}, volume~2.

\bibitem[{Ho and Nvasconcelos(2020)}]{ho2020contrastive}
Chih-Hui Ho and Nuno Nvasconcelos. 2020.
\newblock \href
  {https://proceedings.neurips.cc/paper/2020/file/c68c9c8258ea7d85472dd6fd0015f047-Paper.pdf}
  {Contrastive learning with adversarial examples}.
\newblock In \emph{Advances in Neural Information Processing Systems},
  volume~33.

\bibitem[{Huang et~al.(2019)Huang, Stanforth, Welbl, Dyer, Yogatama, Gowal,
  Dvijotham, and Kohli}]{huang-etal-2019-achieving}
Po-Sen Huang, Robert Stanforth, Johannes Welbl, Chris Dyer, Dani Yogatama, Sven
  Gowal, Krishnamurthy Dvijotham, and Pushmeet Kohli. 2019.
\newblock \href {https://doi.org/10.18653/v1/D19-1419} {Achieving verified
  robustness to symbol substitutions via interval bound propagation}.
\newblock In \emph{Proceedings of the 2019 Conference on Empirical Methods in
  Natural Language Processing and the 9th International Joint Conference on
  Natural Language Processing (EMNLP-IJCNLP)}.

\bibitem[{Iyer et~al.(2017)Iyer, Dandekar, and Csernai}]{qqp}
Shankar Iyer, Nikhil Dandekar, and Kornél Csernai. 2017.
\newblock First quora dataset release: Question pairs.
\newblock
  \url{https://www.quora.com/q/quoradata/First-Quora-Dataset-Release-Question-Pairs}.

\bibitem[{Jacobsen et~al.(2019{\natexlab{a}})Jacobsen, Behrmann, Carlini,
  Tram{\`e}r, and Papernot}]{jacobsen2019safeml}
Joern-Henrik Jacobsen, Jens Behrmann, Nicholas Carlini, Florian Tram{\`e}r, and
  Nicolas Papernot. 2019{\natexlab{a}}.
\newblock Exploiting excessive invariance caused by norm-bounded adversarial
  robustness.
\newblock \emph{Safe Machine Learning workshop at {ICLR}}.

\bibitem[{Jacobsen et~al.(2019{\natexlab{b}})Jacobsen, Behrmann, Zemel, and
  Bethge}]{jacobsen2018excessive}
Joern-Henrik Jacobsen, Jens Behrmann, Richard Zemel, and Matthias Bethge.
  2019{\natexlab{b}}.
\newblock \href {https://openreview.net/forum?id=BkfbpsAcF7} {Excessive
  invariance causes adversarial vulnerability}.
\newblock In \emph{International Conference on Learning Representations}.

\bibitem[{Jia et~al.(2019)Jia, Raghunathan, G{\"o}ksel, and
  Liang}]{jia-etal-2019-certified}
Robin Jia, Aditi Raghunathan, Kerem G{\"o}ksel, and Percy Liang. 2019.
\newblock \href {https://doi.org/10.18653/v1/D19-1423} {Certified robustness to
  adversarial word substitutions}.
\newblock In \emph{Proceedings of the 2019 Conference on Empirical Methods in
  Natural Language Processing and the 9th International Joint Conference on
  Natural Language Processing (EMNLP-IJCNLP)}.

\bibitem[{Jin et~al.(2020)Jin, Jin, Zhou, and
  Szolovits}]{Jin_Jin_Zhou_Szolovits_2020}
Di~Jin, Zhijing Jin, Joey~Tianyi Zhou, and Peter Szolovits. 2020.
\newblock \href {https://doi.org/10.1609/aaai.v34i05.6311} {Is bert really
  robust? a strong baseline for natural language attack on text classification
  and entailment}.
\newblock \emph{Proceedings of the AAAI Conference on Artificial Intelligence},
  34(05).

\bibitem[{Kim et~al.(2020)Kim, Tack, and Hwang}]{Kim2020adversarial}
Minseon Kim, Jihoon Tack, and Sung~Ju Hwang. 2020.
\newblock \href
  {https://proceedings.neurips.cc/paper/2020/file/1f1baa5b8edac74eb4eaa329f14a0361-Paper.pdf}
  {Adversarial self-supervised contrastive learning}.
\newblock In \emph{Advances in Neural Information Processing Systems},
  volume~33.

\bibitem[{Kumar and Boulanger(2020)}]{Kumar2020ExplainableAE}
Vivekanandan~S. Kumar and David Boulanger. 2020.
\newblock Explainable automated essay scoring: Deep learning really has
  pedagogical value.
\newblock In \emph{Frontiers in Education}.

\bibitem[{Li et~al.(2021)Li, Zhang, Peng, Chen, Brockett, Sun, and
  Dolan}]{li-etal-2021-contextualized}
Dianqi Li, Yizhe Zhang, Hao Peng, Liqun Chen, Chris Brockett, Ming-Ting Sun,
  and Bill Dolan. 2021.
\newblock \href {https://doi.org/10.18653/v1/2021.naacl-main.400}
  {Contextualized perturbation for textual adversarial attack}.
\newblock In \emph{Proceedings of the 2021 Conference of the North American
  Chapter of the Association for Computational Linguistics: Human Language
  Technologies}.

\bibitem[{Li et~al.(2020)Li, Ma, Guo, Xue, and Qiu}]{li-etal-2020-bert-attack}
Linyang Li, Ruotian Ma, Qipeng Guo, Xiangyang Xue, and Xipeng Qiu. 2020.
\newblock \href {https://doi.org/10.18653/v1/2020.emnlp-main.500}
  {{BERT}-{ATTACK}: Adversarial attack against {BERT} using {BERT}}.
\newblock In \emph{Proceedings of the 2020 Conference on Empirical Methods in
  Natural Language Processing (EMNLP)}.

\bibitem[{Liu et~al.(2019)Liu, Ott, Goyal, Du, Joshi, Chen, Levy, Lewis,
  Zettlemoyer, and Stoyanov}]{liu2019roberta}
Yinhan Liu, Myle Ott, Naman Goyal, Jingfei Du, Mandar Joshi, Danqi Chen, Omer
  Levy, Mike Lewis, Luke Zettlemoyer, and Veselin Stoyanov. 2019.
\newblock \href {http://arxiv.org/abs/1907.11692} {{RoBERTa}: A robustly
  optimized {BERT} pretraining approach}.
\newblock \emph{CoRR}, abs/1907.11692.

\bibitem[{Madry et~al.(2018)Madry, Makelov, Schmidt, Tsipras, and
  Vladu}]{madry2018towards}
Aleksander Madry, Aleksandar Makelov, Ludwig Schmidt, Dimitris Tsipras, and
  Adrian Vladu. 2018.
\newblock \href {https://openreview.net/forum?id=rJzIBfZAb} {Towards deep
  learning models resistant to adversarial attacks}.
\newblock In \emph{International Conference on Learning Representations}.

\bibitem[{Miller(1995)}]{miller-wordnet-1995}
George~A. Miller. 1995.
\newblock \href {https://doi.org/10.1145/219717.219748} {{WordNet}: A lexical
  database for {E}nglish}.
\newblock \emph{Communications of the ACM}, 38(11).

\bibitem[{Morris et~al.(2020)Morris, Lifland, Lanchantin, Ji, and
  Qi}]{morris-etal-2020-reevaluating}
John Morris, Eli Lifland, Jack Lanchantin, Yangfeng Ji, and Yanjun Qi. 2020.
\newblock \href {https://doi.org/10.18653/v1/2020.findings-emnlp.341}
  {Reevaluating adversarial examples in natural language}.
\newblock In \emph{Findings of the Association for Computational Linguistics:
  EMNLP 2020}.

\bibitem[{Niu and Bansal(2018)}]{niu-bansal-2018-adversarial}
Tong Niu and Mohit Bansal. 2018.
\newblock \href {https://doi.org/10.18653/v1/K18-1047} {Adversarial
  over-sensitivity and over-stability strategies for dialogue models}.
\newblock In \emph{Proceedings of the 22nd Conference on Computational Natural
  Language Learning}.

\bibitem[{Pennington et~al.(2014)Pennington, Socher, and
  Manning}]{pennington-etal-2014-glove}
Jeffrey Pennington, Richard Socher, and Christopher Manning. 2014.
\newblock \href {https://doi.org/10.3115/v1/D14-1162} {{G}lo{V}e: Global
  vectors for word representation}.
\newblock In \emph{Proceedings of the 2014 Conference on Empirical Methods in
  Natural Language Processing ({EMNLP})}.

\bibitem[{Ren et~al.(2019)Ren, Deng, He, and Che}]{ren-etal-2019-generating}
Shuhuai Ren, Yihe Deng, Kun He, and Wanxiang Che. 2019.
\newblock \href {https://doi.org/10.18653/v1/P19-1103} {Generating natural
  language adversarial examples through probability weighted word saliency}.
\newblock In \emph{Proceedings of the 57th Annual Meeting of the Association
  for Computational Linguistics}.

\bibitem[{Schroff et~al.(2015)Schroff, Kalenichenko, and
  Philbin}]{schroff2015facenet}
Florian Schroff, Dmitry Kalenichenko, and James Philbin. 2015.
\newblock \href {https://doi.org/10.1109/CVPR.2015.7298682} {{FaceNet}: A
  unified embedding for face recognition and clustering}.
\newblock In \emph{2015 IEEE Conference on Computer Vision and Pattern
  Recognition (CVPR)}.

\bibitem[{Tramer et~al.(2020)Tramer, Behrmann, Carlini, Papernot, and
  Jacobsen}]{pmlr-v119-tramer20a}
Florian Tramer, Jens Behrmann, Nicholas Carlini, Nicolas Papernot, and
  Joern-Henrik Jacobsen. 2020.
\newblock \href {https://proceedings.mlr.press/v119/tramer20a.html}
  {Fundamental tradeoffs between invariance and sensitivity to adversarial
  perturbations}.
\newblock In \emph{Proceedings of the 37th International Conference on Machine
  Learning}, volume 119 of \emph{Proceedings of Machine Learning Research}.

\bibitem[{van~der Maaten and Hinton(2008)}]{tsne-vandermaaten08a}
Laurens van~der Maaten and Geoffrey Hinton. 2008.
\newblock \href {http://jmlr.org/papers/v9/vandermaaten08a.html} {Visualizing
  data using {t-SNE}}.
\newblock \emph{Journal of Machine Learning Research}, 9(86).

\bibitem[{Wang et~al.(2018)Wang, Singh, Michael, Hill, Levy, and
  Bowman}]{wang-etal-2018-glue}
Alex Wang, Amanpreet Singh, Julian Michael, Felix Hill, Omer Levy, and Samuel
  Bowman. 2018.
\newblock \href {https://doi.org/10.18653/v1/W18-5446} {{GLUE}: A multi-task
  benchmark and analysis platform for natural language understanding}.
\newblock In \emph{Proceedings of the 2018 {EMNLP} Workshop {B}lackbox{NLP}:
  Analyzing and Interpreting Neural Networks for {NLP}}.

\bibitem[{Welbl et~al.(2020)Welbl, Minervini, Bartolo, Stenetorp, and
  Riedel}]{welbl-etal-2020-undersensitivity}
Johannes Welbl, Pasquale Minervini, Max Bartolo, Pontus Stenetorp, and
  Sebastian Riedel. 2020.
\newblock \href {https://doi.org/10.18653/v1/2020.findings-emnlp.103}
  {Undersensitivity in neural reading comprehension}.
\newblock In \emph{Findings of the Association for Computational Linguistics:
  EMNLP 2020}.

\bibitem[{Williams et~al.(2018)Williams, Nangia, and
  Bowman}]{williams-etal-2018-broad}
Adina Williams, Nikita Nangia, and Samuel Bowman. 2018.
\newblock \href {https://doi.org/10.18653/v1/N18-1101} {A broad-coverage
  challenge corpus for sentence understanding through inference}.
\newblock In \emph{Proceedings of the 2018 Conference of the North {A}merican
  Chapter of the Association for Computational Linguistics: Human Language
  Technologies, Volume 1 (Long Papers)}.

\bibitem[{Wulczyn et~al.(2017)Wulczyn, Thain, and Dixon}]{wiki-toxic}
Ellery Wulczyn, Nithum Thain, and Lucas Dixon. 2017.
\newblock \href {https://doi.org/10.1145/3038912.3052591} {Ex {M}achina:
  Personal attacks seen at scale}.
\newblock In \emph{Proceedings of the 26th International Conference on World
  Wide Web}.

\bibitem[{Ye et~al.(2020)Ye, Gong, and Liu}]{ye-etal-2020-safer}
Mao Ye, Chengyue Gong, and Qiang Liu. 2020.
\newblock \href {https://doi.org/10.18653/v1/2020.acl-main.317} {{SAFER}: A
  structure-free approach for certified robustness to adversarial word
  substitutions}.
\newblock In \emph{Proceedings of the 58th Annual Meeting of the Association
  for Computational Linguistics}.

\bibitem[{Yoo and Qi(2021)}]{yoo-qi-2021-towards-improving}
Jin~Yong Yoo and Yanjun Qi. 2021.
\newblock \href {https://aclanthology.org/2021.findings-emnlp.81} {Towards
  improving adversarial training of {NLP} models}.
\newblock In \emph{Findings of the Association for Computational Linguistics:
  EMNLP 2021}.

\end{thebibliography}
\bibliographystyle{acl_natbib}

\clearpage
\appendix

\onecolumn
\label{sec:appendix}

\section{Fickleness and Obstinacy Robustness Tradeoffs}

\subsection{Synonym and Antonym Attack Robustness Tradeoffs}
\label{app:qqp-tradeoff}

\begin{figure}[!h]
    \centering
    \begin{subfigure}[b]{0.45\linewidth}
         \centering
         \includegraphics[width=1.2\textwidth]{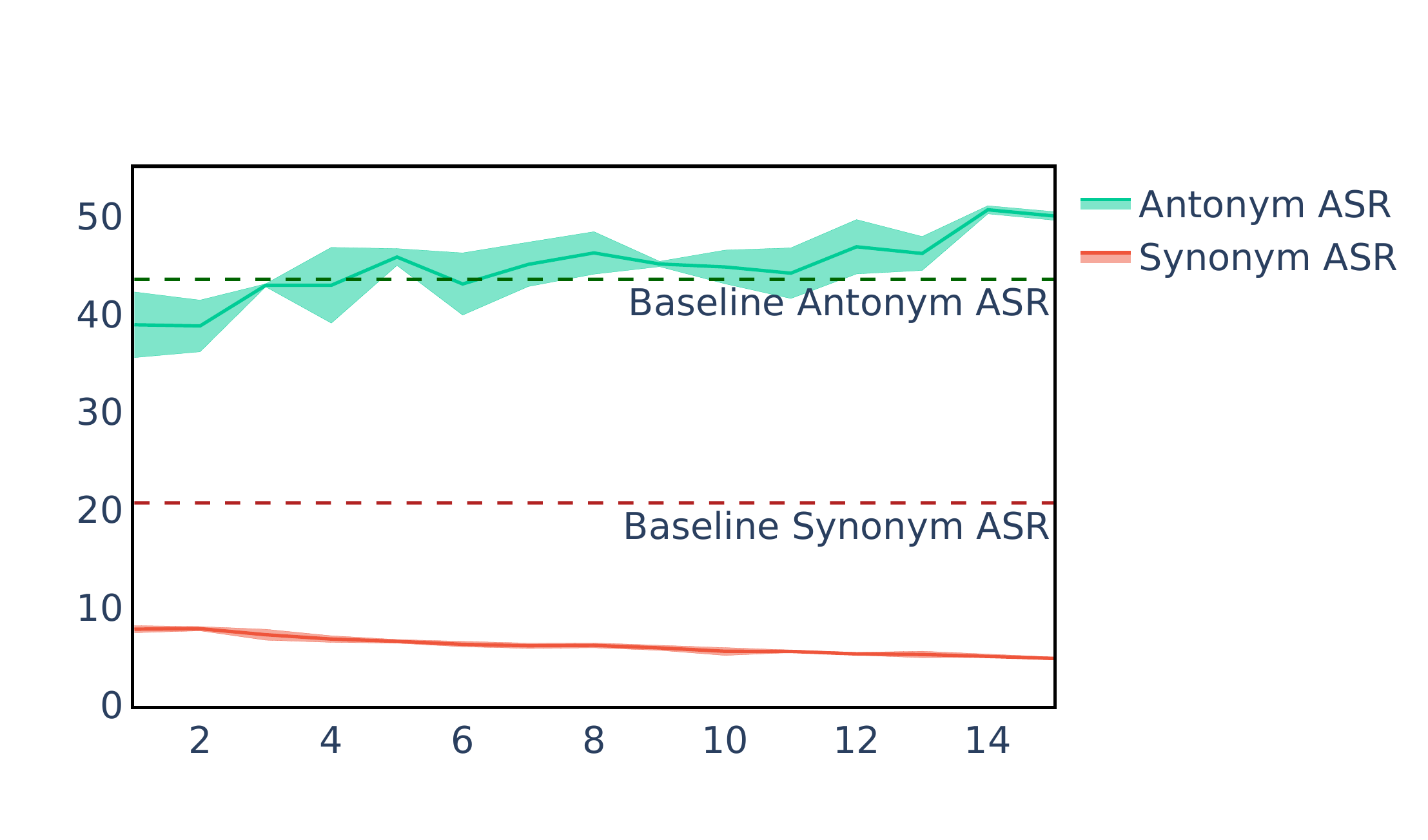}
         \caption{QQP (BERT)}
         \label{fig:bert-qqp-tradeoff}
     \end{subfigure}
     \hfill
     \begin{subfigure}[b]{0.45\linewidth}
         \centering
         \includegraphics[width=1.0\textwidth]{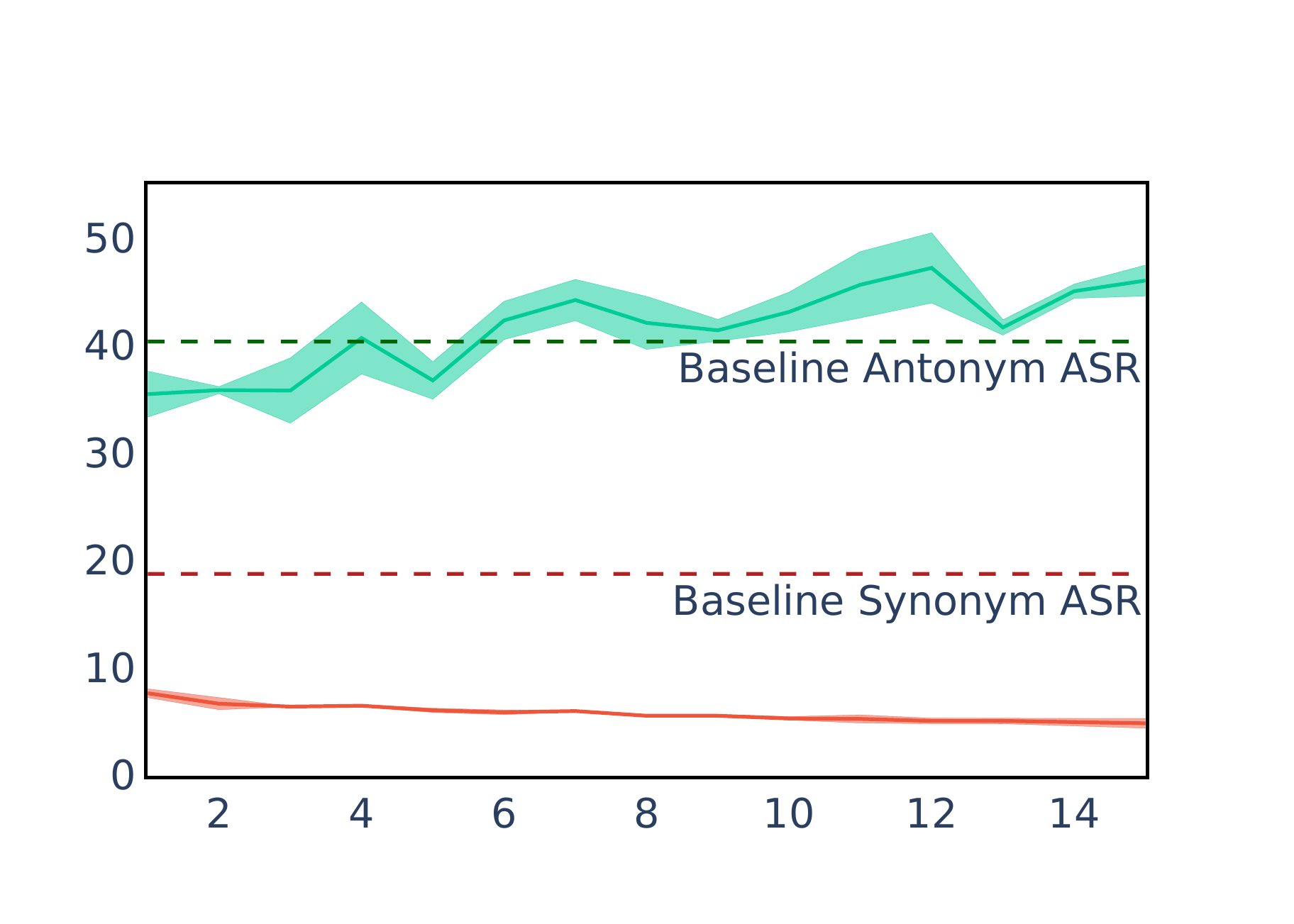}
         \caption{QQP (RoBERTa)}
         \label{fig:roberta-qqp-tradeoff}
     \end{subfigure}
     \begin{subfigure}[b]{0.45\linewidth}
         \centering
         \includegraphics[width=1.0\textwidth]{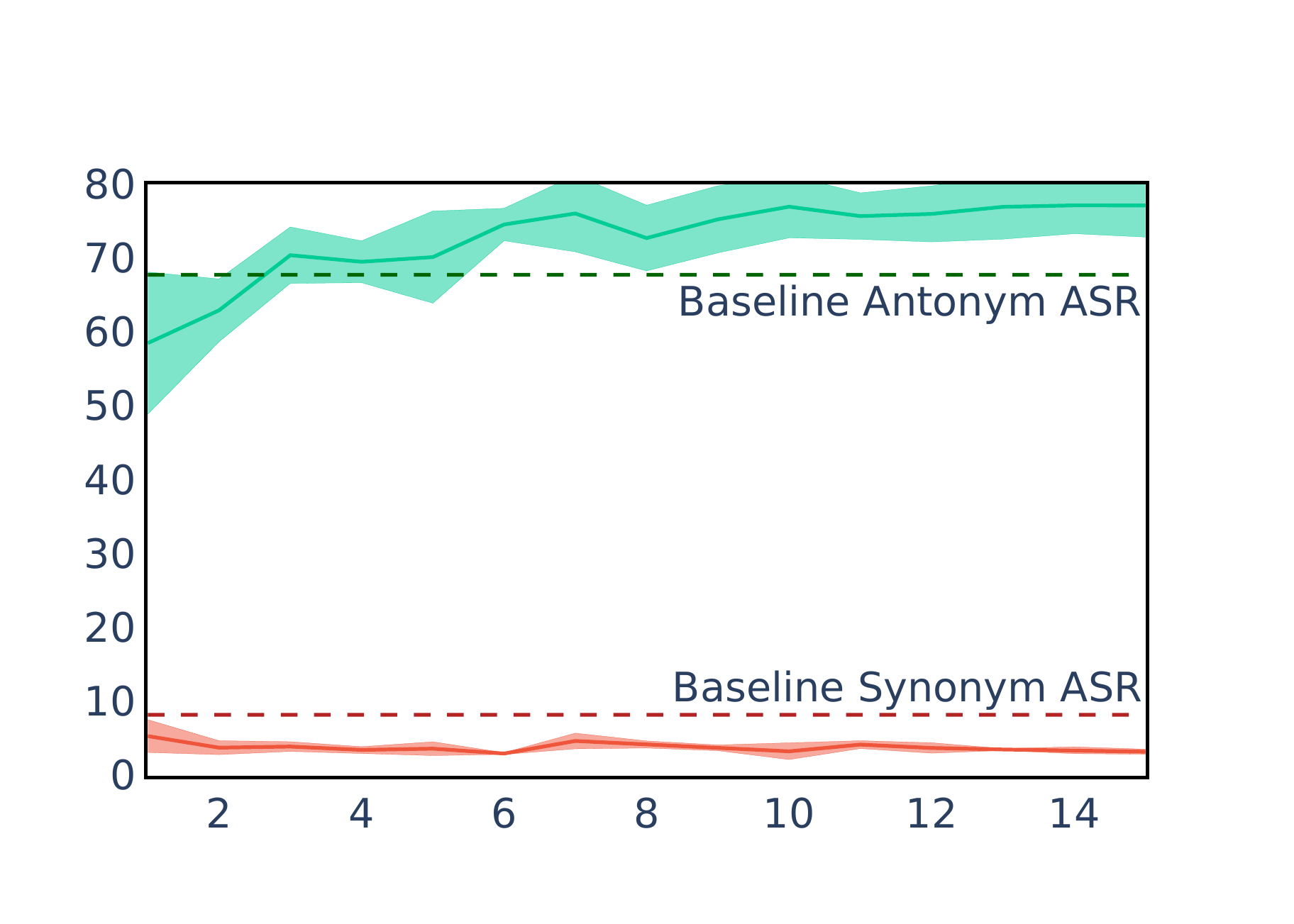}
         \caption{MRPC (BERT)}
         \label{fig:bert-mrpc-tradeoff}
     \end{subfigure}
     \hfill
     \begin{subfigure}[b]{0.45\linewidth}
         \centering
         \includegraphics[width=1\textwidth]{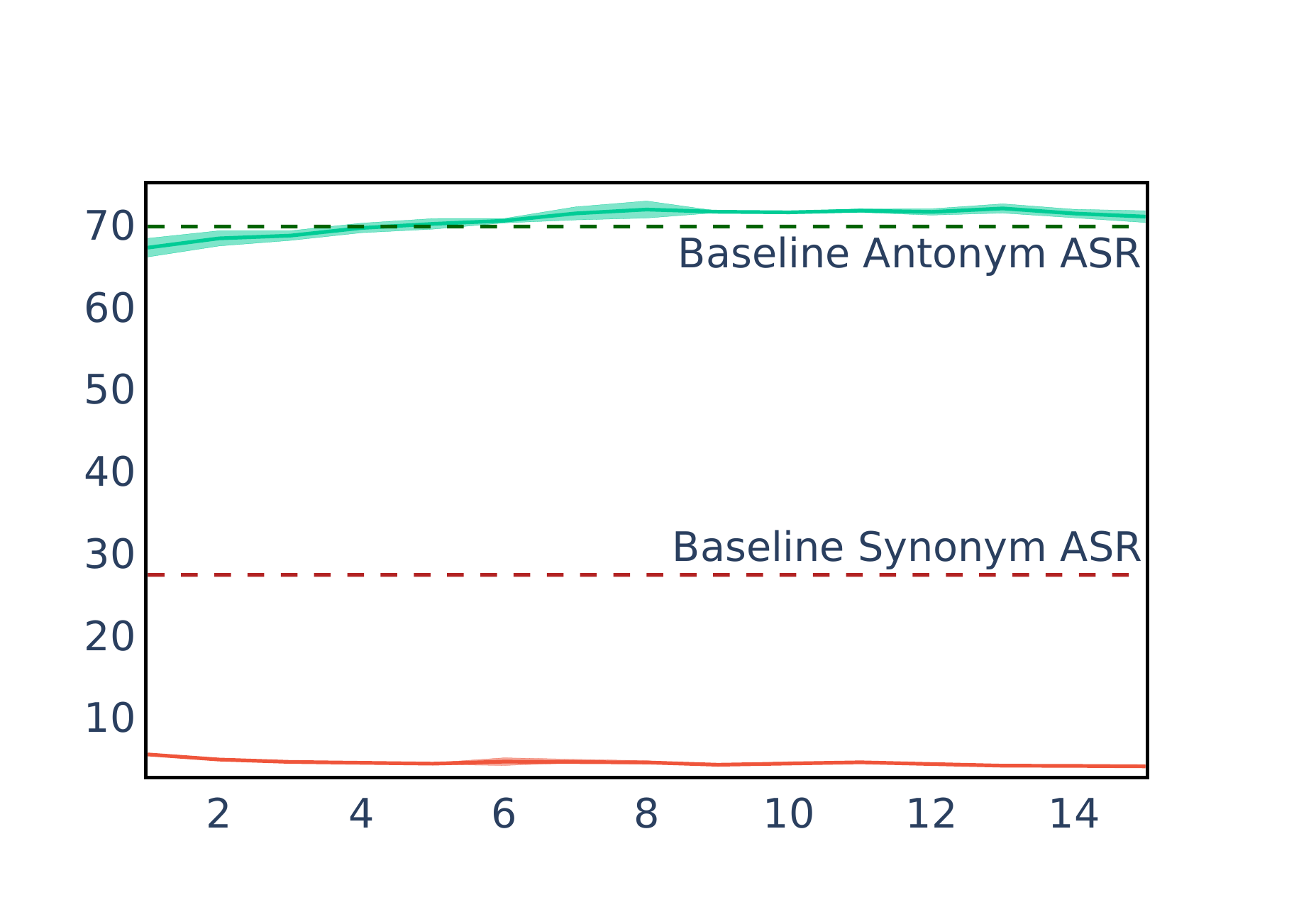}
         \caption{SNLI (BERT)}
         \label{fig:bert-snli-tradeoff}
     \end{subfigure}
    \caption{Robustness tradeoffs between synonym and antonym based attacks on QQP and MRPC dataset. The figure shows the average and standard deviation across 3 different runs.}
    \label{fig:tradeoff-qqp}
\end{figure}

\clearpage

\subsection{Synonym and Negation Attack Robustness Tradeoffs}

\begin{figure*}[!h]
     \centering
     \begin{subfigure}[b]{0.45\linewidth}
         \centering
         \includegraphics[width=1.2\textwidth]{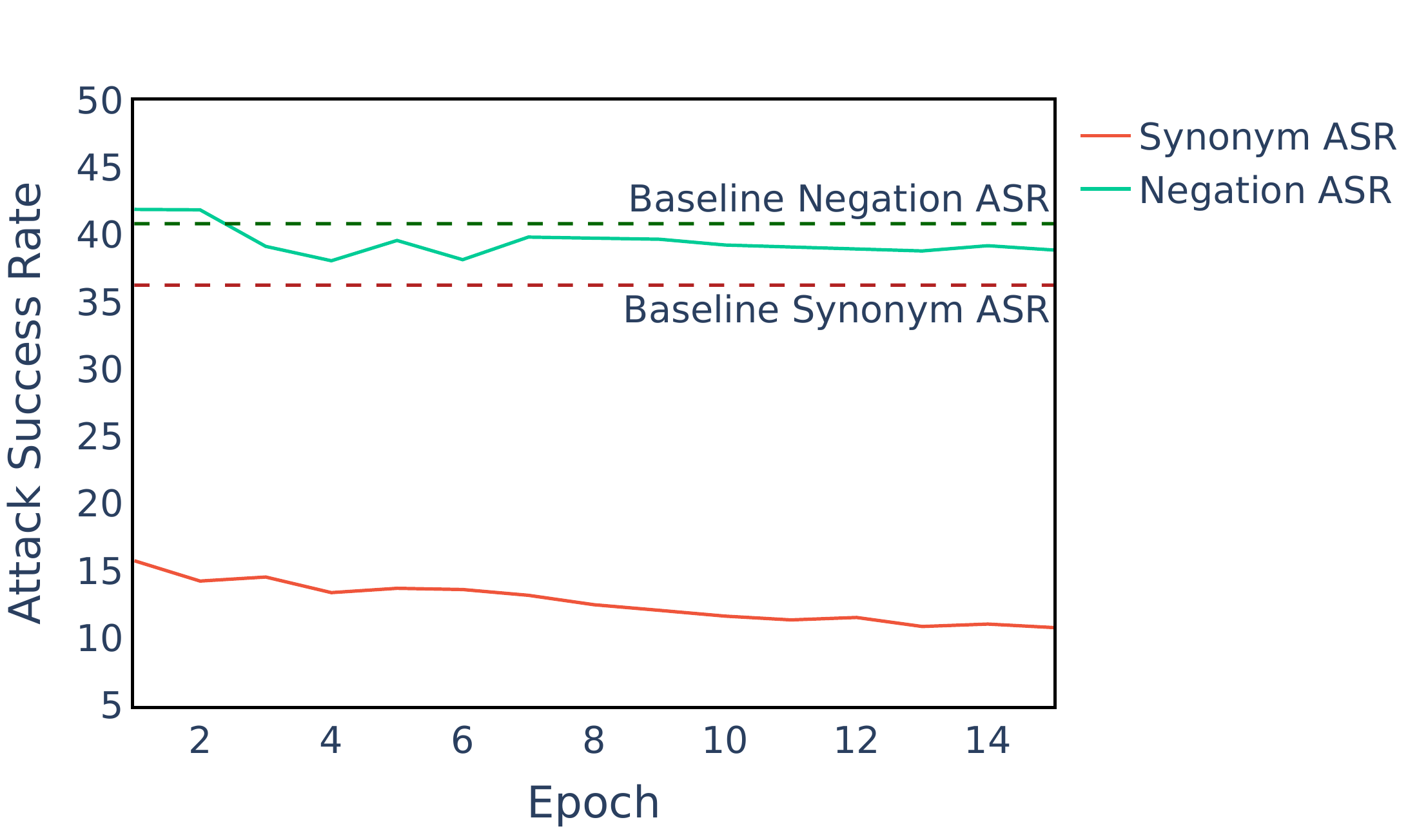}
         \caption{MNLI (BERT)}
         \label{fig:bert-mnli-negation-tradeoff}
     \end{subfigure}
     \hfill
     \begin{subfigure}[b]{0.45\linewidth}
         \centering
         \includegraphics[width=1.05\textwidth]{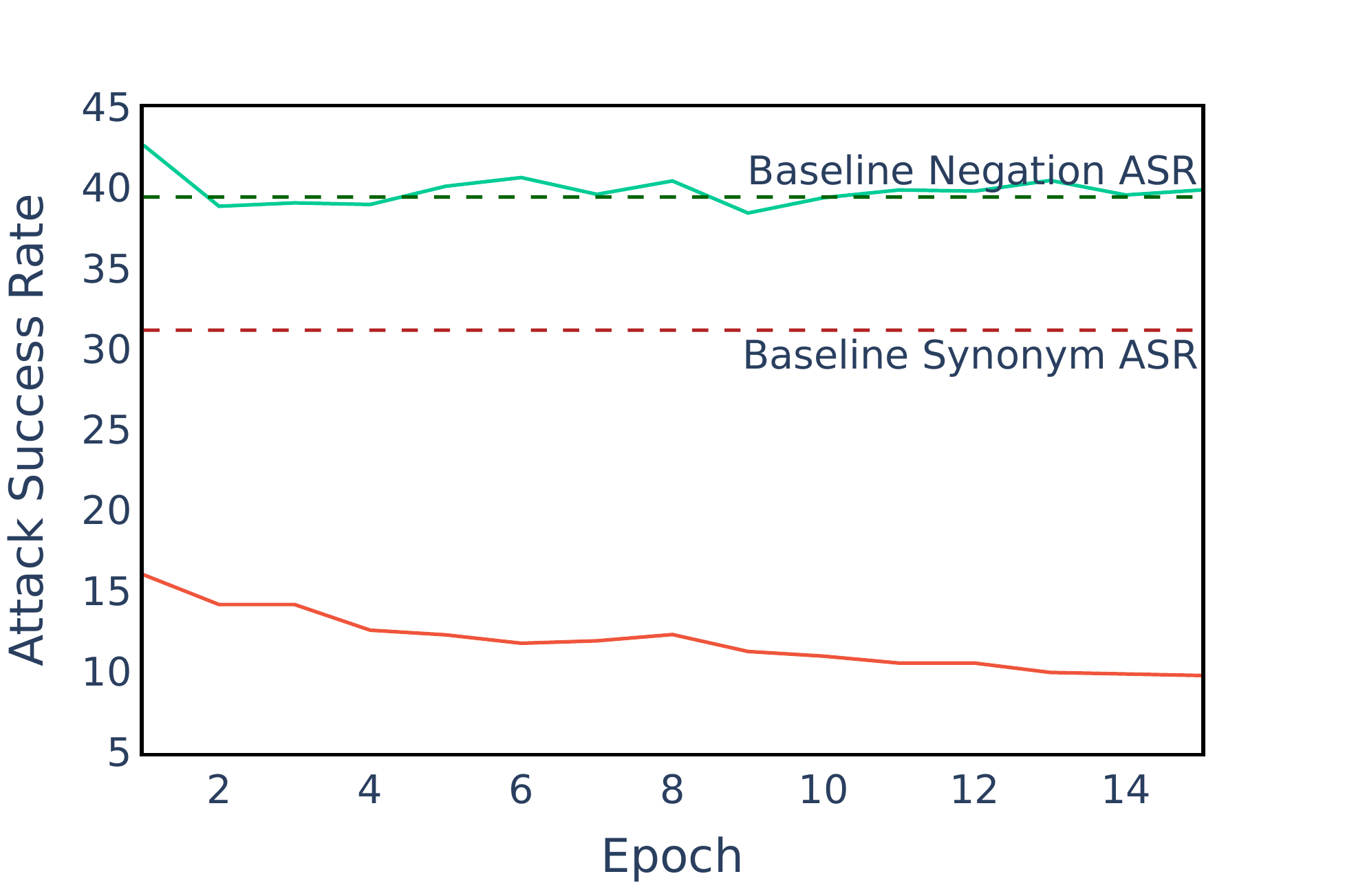}
         \caption{MNLI (RoBERTa)}
         \label{fig:roberta-mnli-negation-tradeoff}
     \end{subfigure}
     \begin{subfigure}[b]{0.45\linewidth}
         \centering
         \includegraphics[width=1.0\textwidth]{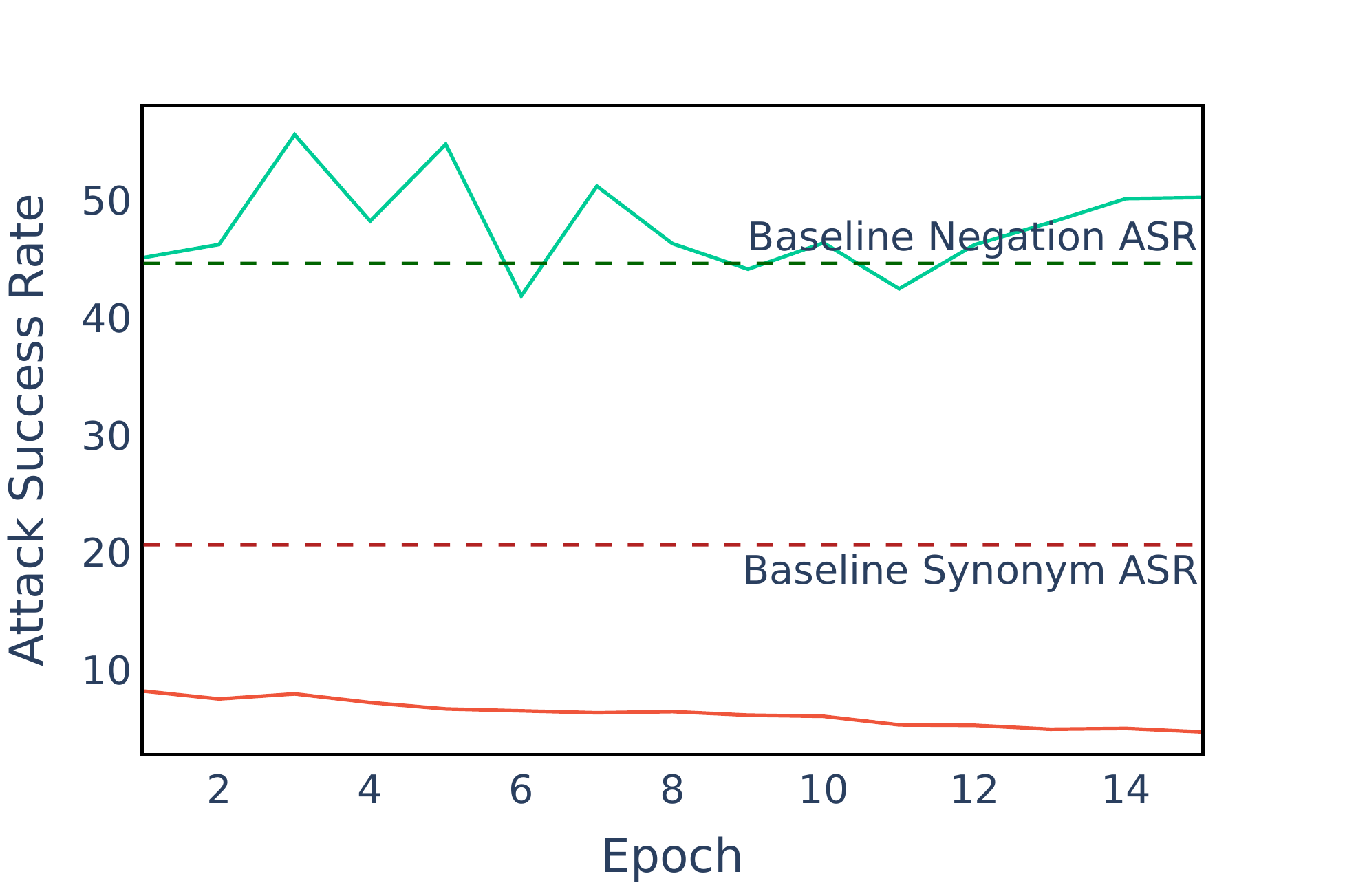}
         \caption{QQP (BERT)}
         \label{fig:bert-qqp-negation-tradeoff}
     \end{subfigure}
     \hfill
     \begin{subfigure}[b]{0.45\linewidth}
         \centering
         \includegraphics[width=1.0\textwidth]{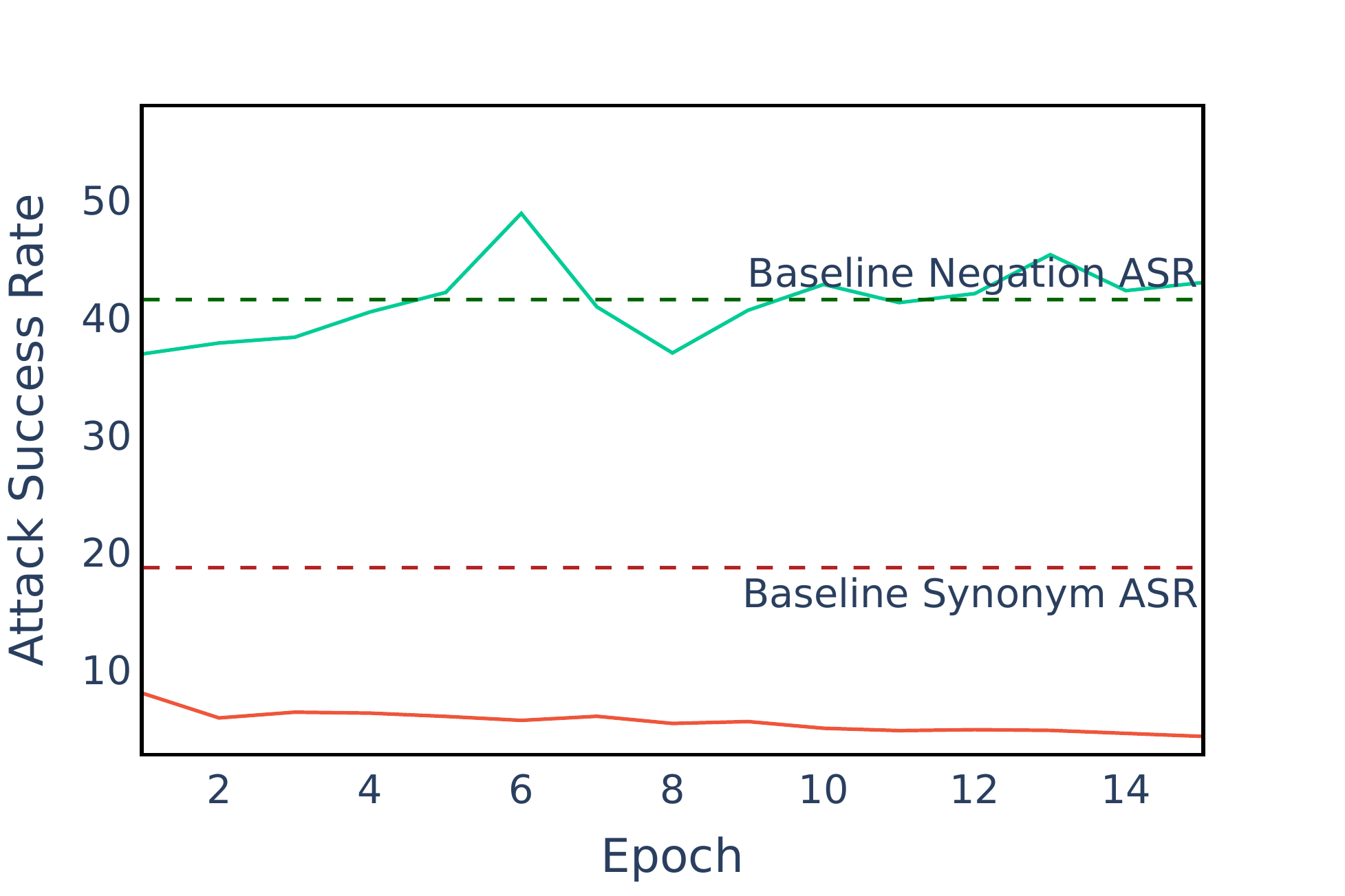}
         \caption{QQP (RoBERTa)}
         \label{fig:roberta-qqp-negation-tradeoff}
     \end{subfigure}
     \caption{Negation attack success rate on models at each epoch when training with SAFER.}
     \label{fig:negation-tradeoff}
\end{figure*}

We test how negation attack success rate would change as the model robustness against synonym attack increases. For negation attack, we add negation to a verb in the sentence, i.e., ``\textit{I can do it}'' to ``\textit{I can't do it}'', or remove negation from a sentence, i.e., ``\textit{I am not going}'' to ``\textit{I am going}''. We follow similar setup as in Section~\ref{sec:tradeoff}. We found that there exists a tradeoff between synonym-based adversarial examples and negation-based adversarial examples on QQP task, but found no significant tradeoff on MNLI task, as shown in Figure~\ref{fig:negation-tradeoff}.

\clearpage

\section{Fickleness Robust Training with Varying Batch Size}
\label{app:safer-varying-batch-size}

\subsection{Synonym and Antonym Robustness Tradeoffs on QQP Task}
\begin{figure*}[!h]
    \centering
    \begin{subfigure}[b]{0.43\linewidth}
         \centering
         \includegraphics[width=1.25\textwidth]{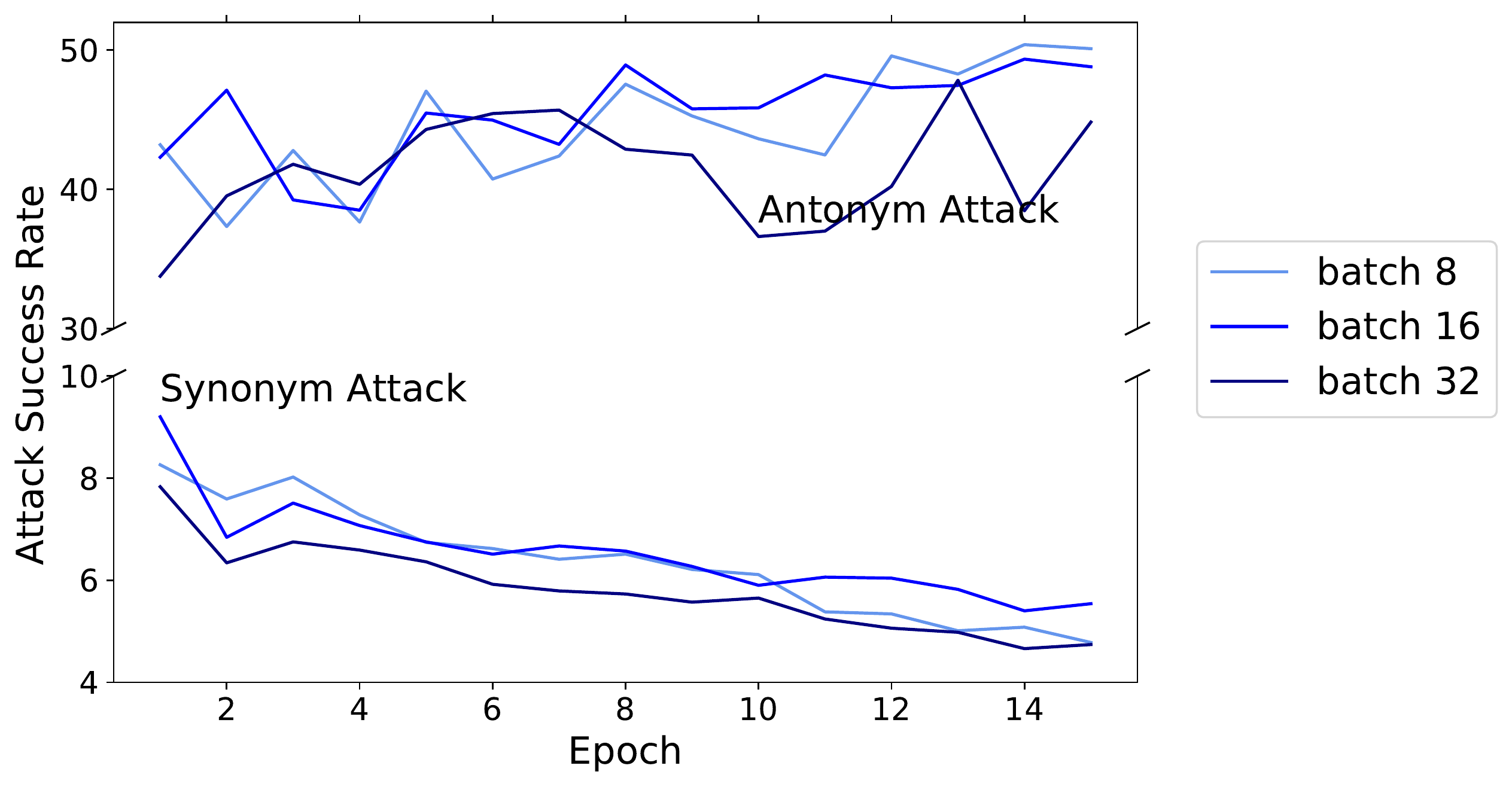}
         \caption{QQP (BERT)}
         \label{fig:bert-qqp-batch-size-asr}
     \end{subfigure}
     \hfill
     \begin{subfigure}[b]{0.43\linewidth}
         \centering
         \includegraphics[width=0.95\textwidth]{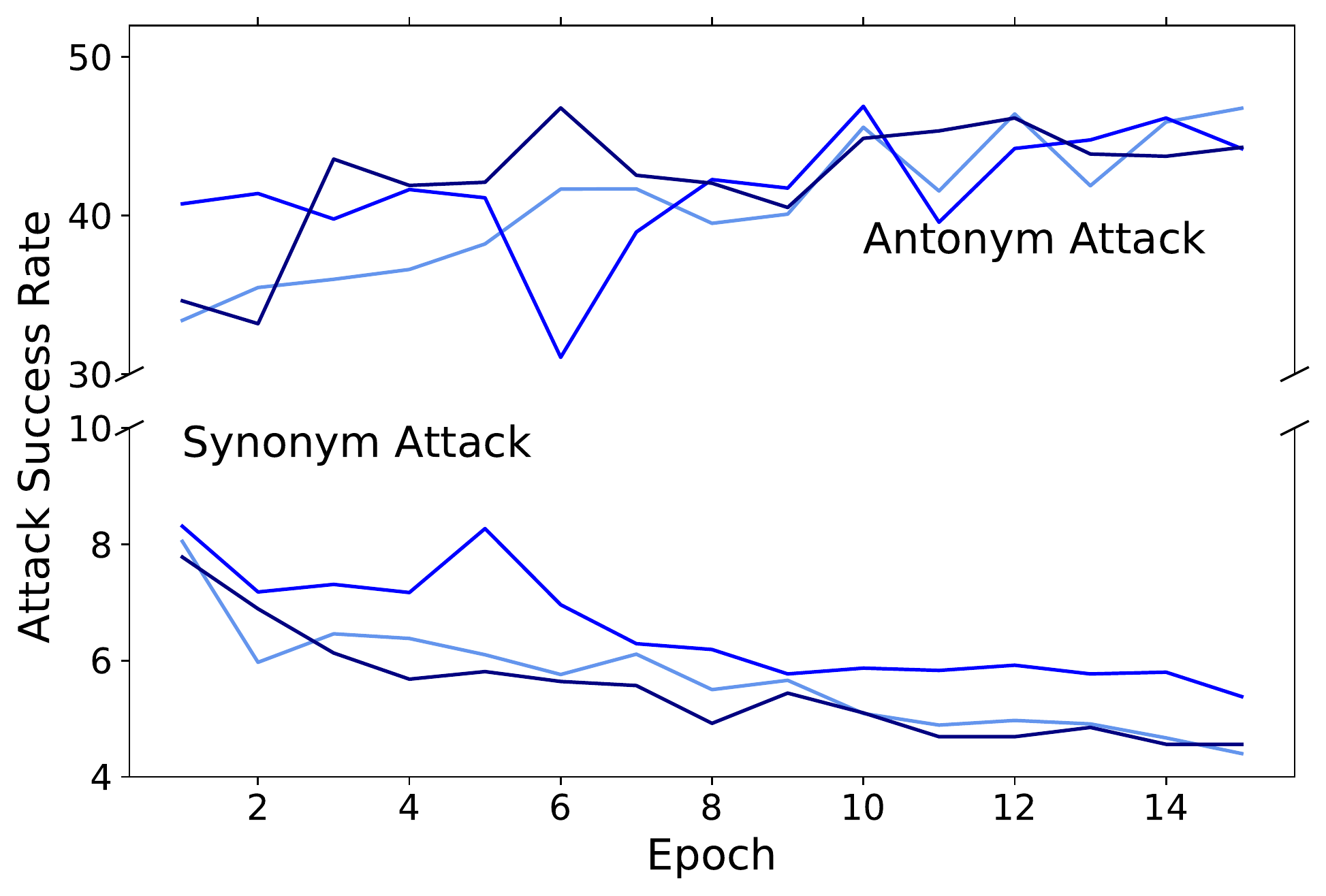}
         \caption{QQP (RoBERTa)}
         \label{fig:roberta-qqp-batch-size-asr}
     \end{subfigure}
    \caption{The synonym and antonym attack success rate at each SAFER training epoch with varying batch size.}
    \label{fig:batch-size-asr-qqp}
\end{figure*}

\subsection{Evaluation Accuracy at Each SAFER Training Epoch}
\begin{figure*}[!h]
    \centering
    \begin{subfigure}[b]{0.48\linewidth}
         \includegraphics[width=1\textwidth]{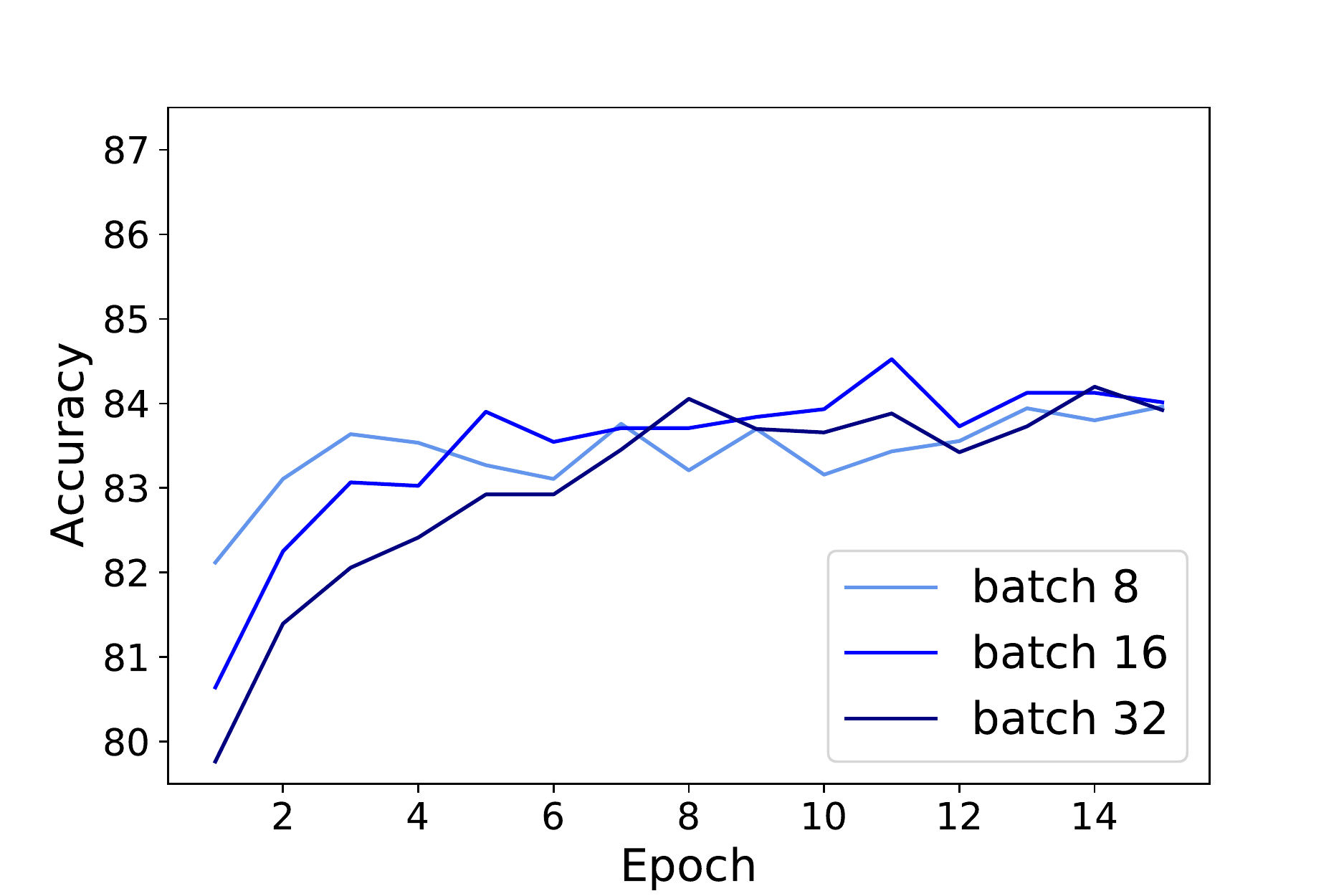}
         \caption{MNLI (BERT)}
     \end{subfigure}
     \hfill
     \begin{subfigure}[b]{0.48\linewidth}
         \includegraphics[width=1\textwidth]{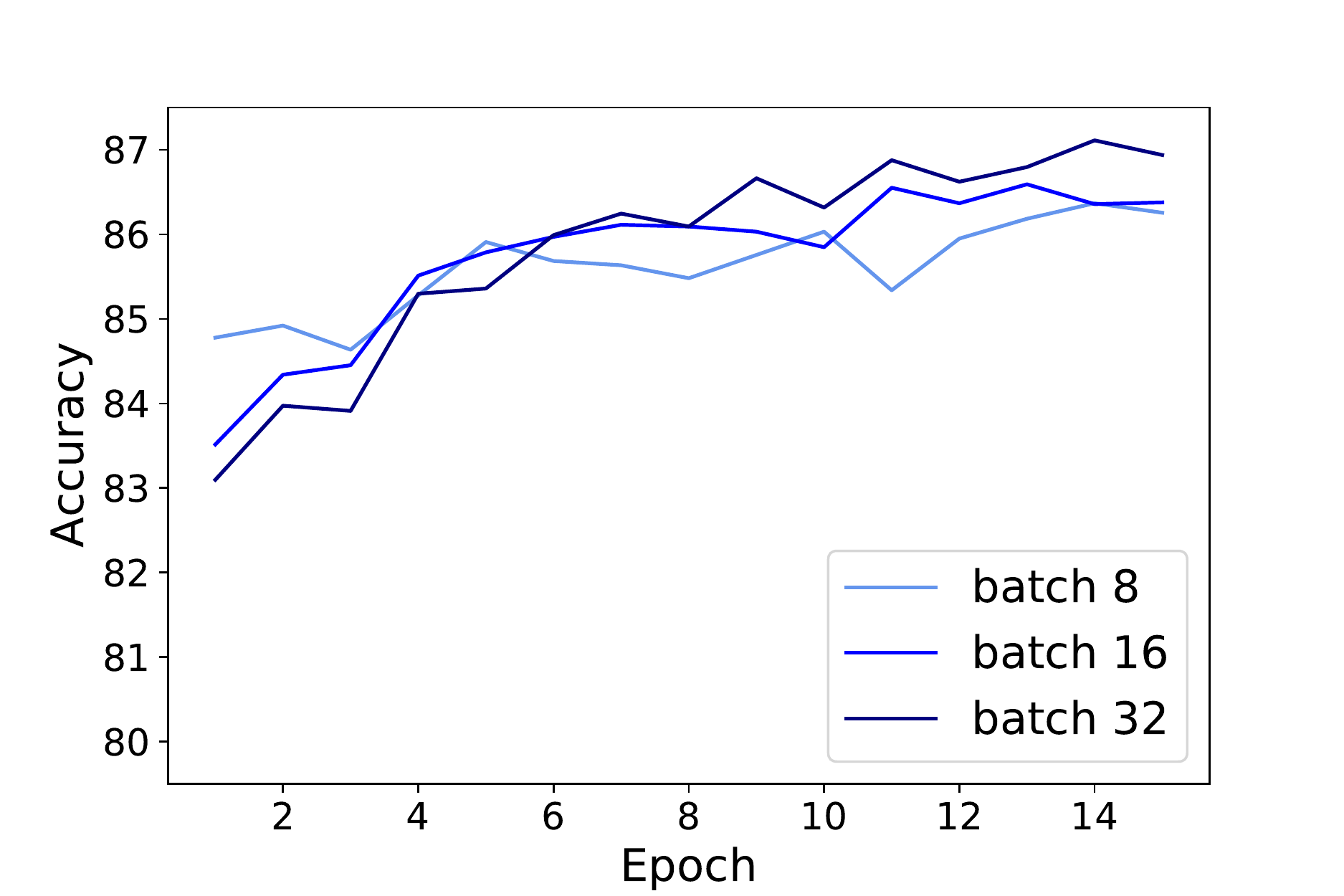}
         \caption{MNLI (RoBERTa)}
     \end{subfigure}
     \begin{subfigure}[b]{0.48\linewidth}
         \includegraphics[width=1\textwidth]{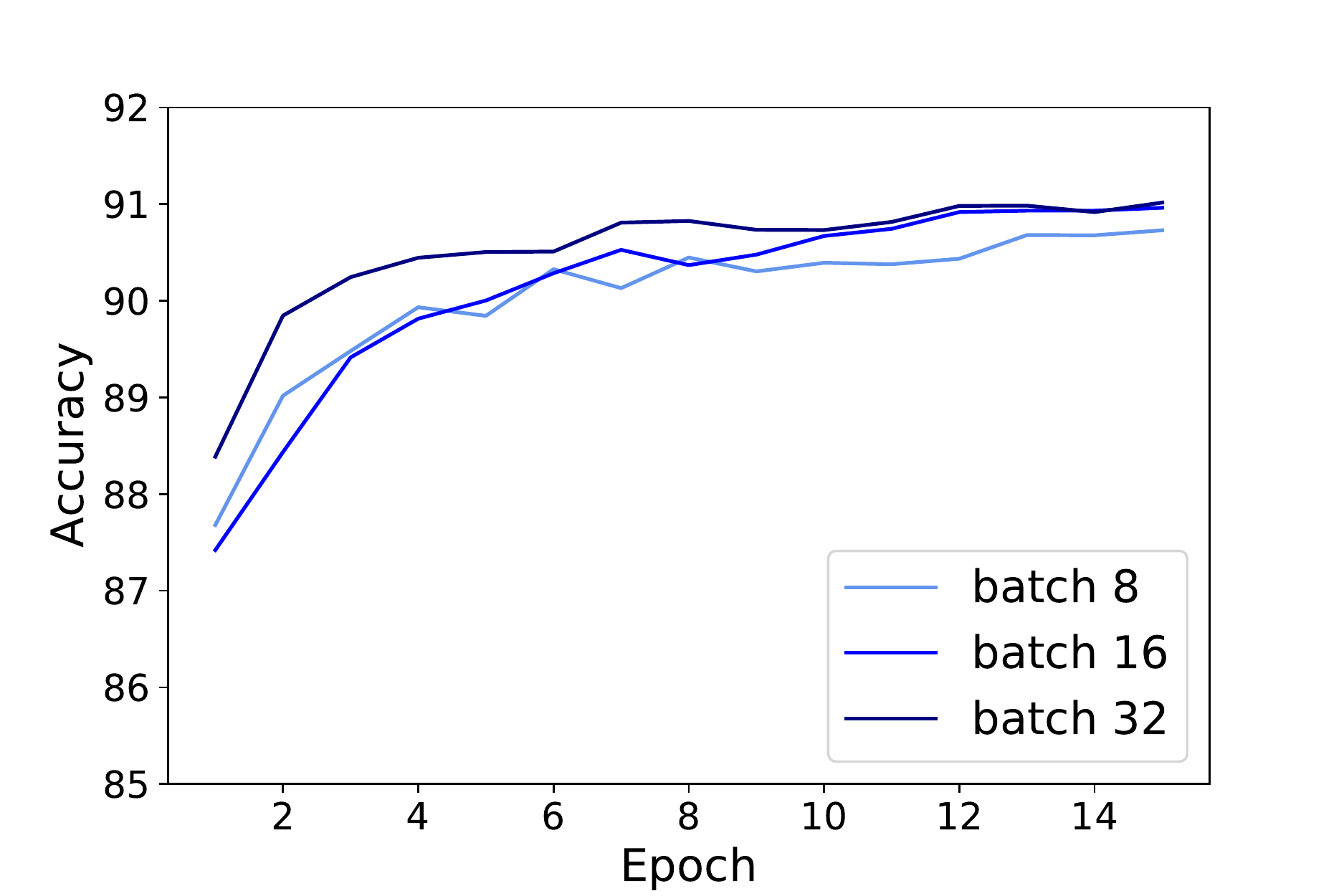}
         \caption{QQP (BERT)}
     \end{subfigure}
     \hfill
     \begin{subfigure}[b]{0.48\linewidth}
         \includegraphics[width=1\textwidth]{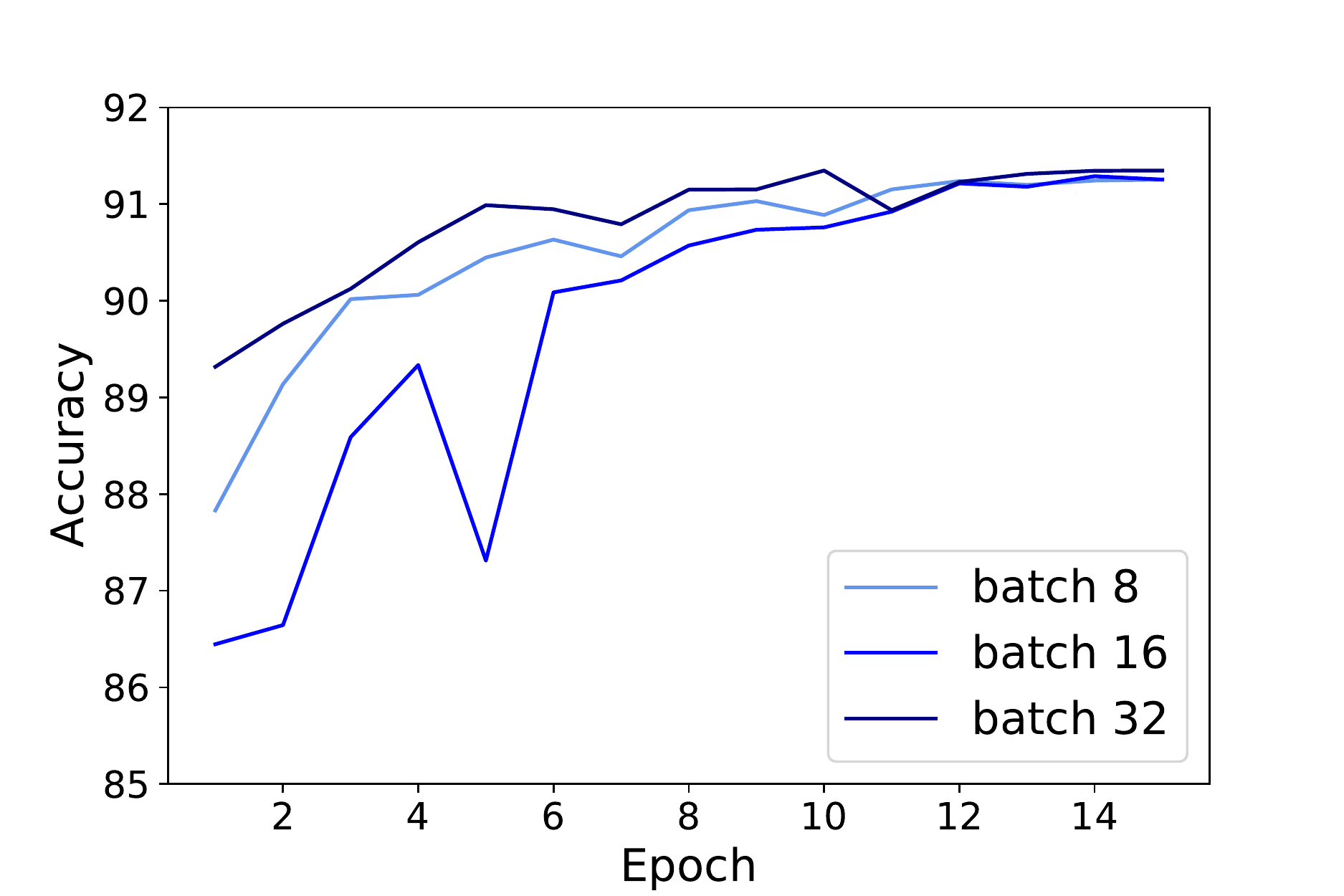}
         \caption{QQP (RoBERTa)}
     \end{subfigure}
    \caption{The evaluation accuracy on original validation set at each SAFER training epoch with varying batch size.}
    \label{fig:batch-size-eval-acc}
\end{figure*}

\clearpage

\section{Balanced Adversarial Training}
\label{app:bat-training-details-and-parameters}

\subsection{Balanced Adversarial Training Details}
\label{app:bat-training-details}
We implement BAT similarly to the SAFER training method as described in Section~\ref{sec:tradeoff-setup} where we randomly perturb the inputs with words from the synonym/antonym substitution sets. 

We train the BERT and RoBERTa models for 2 or 3 epochs with a learning rate of $2 \times 10^{-5}$ or $3 \times 10^{-5}$ and batch size of 32. For contrastive loss weights and margin in BAT-Pairwise and BAT-Triplet, we perform hyperparmeter search and choose the ones with best performance (see Appendix~\ref{app:bat-hyperparameter-search}).

\subsection{BAT Hyperparameter Search}
\label{app:bat-hyperparameter-search}

\subsubsection{Contrastive Loss Weights}
\begin{figure*}[!h]
     \centering
     \begin{subfigure}[b]{0.46\linewidth}
         \includegraphics[width=1.15\textwidth]{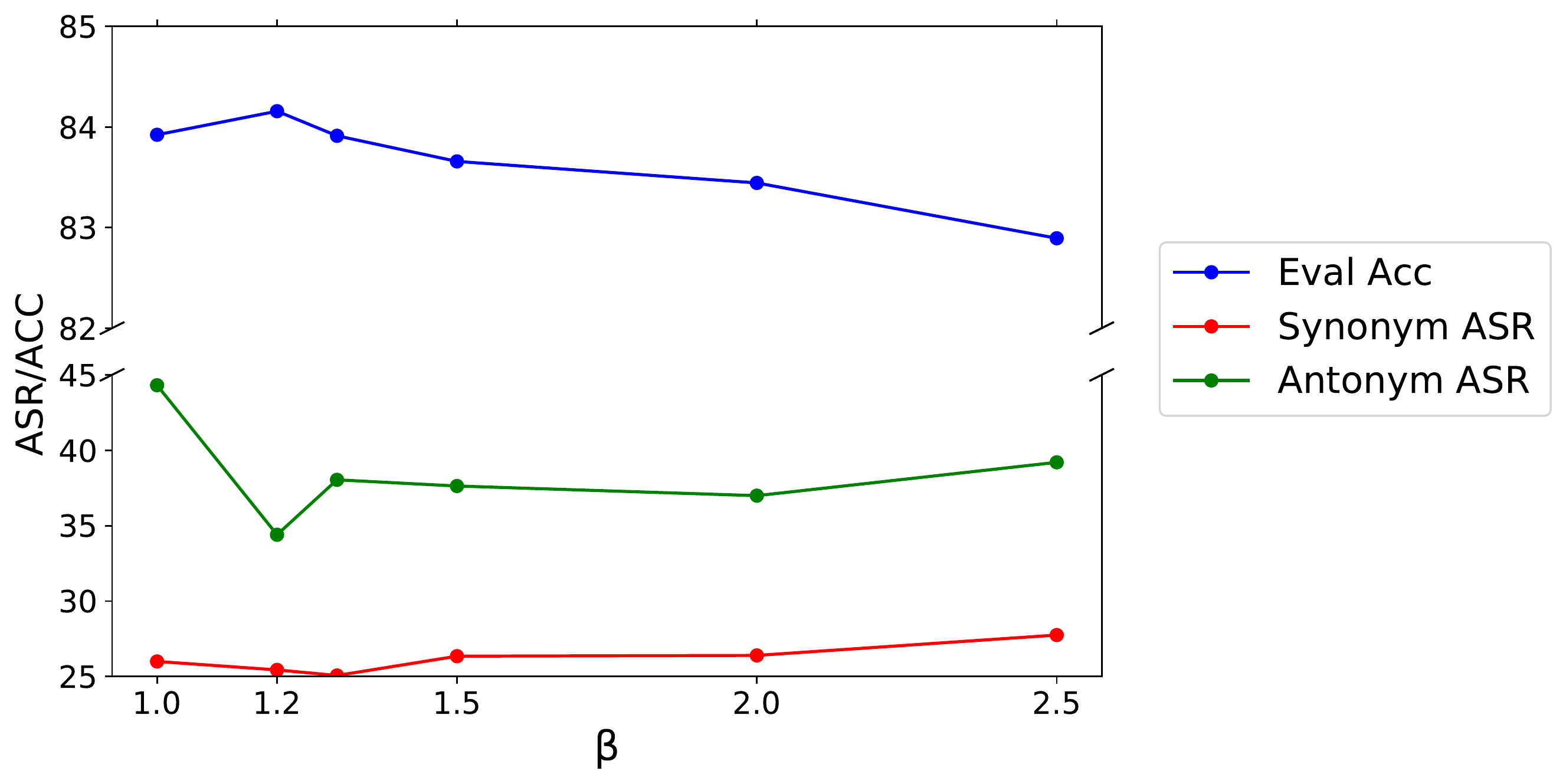}
         \caption{Varying $\beta$ with $\alpha=1.0$}
     \end{subfigure}
     \hfill
     \begin{subfigure}[b]{0.46\linewidth}
         \includegraphics[width=0.85\textwidth]{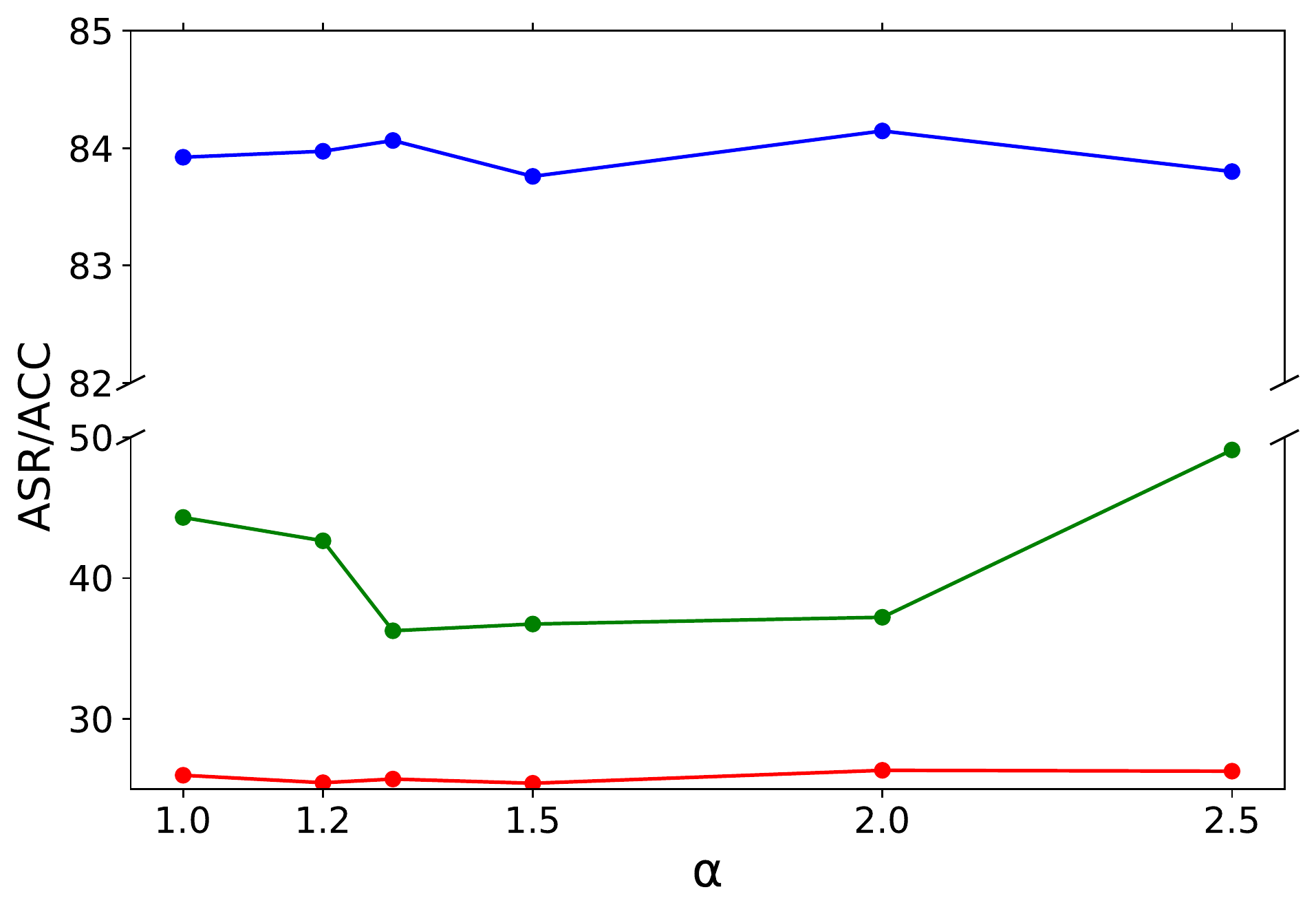}
         \caption{Varying $\alpha$ with $\beta=1.0$}
     \end{subfigure}
     \caption{Performance of BERT models trained on MNLI tasks with different fickle ($\alpha$) and obstinate ($\beta$) loss weights in BAT-Pairwise.}
     \label{fig:BAT-pairwise-weights}
\end{figure*}

\begin{figure*}[!h]
     \centering
     \includegraphics[width=0.6\linewidth]{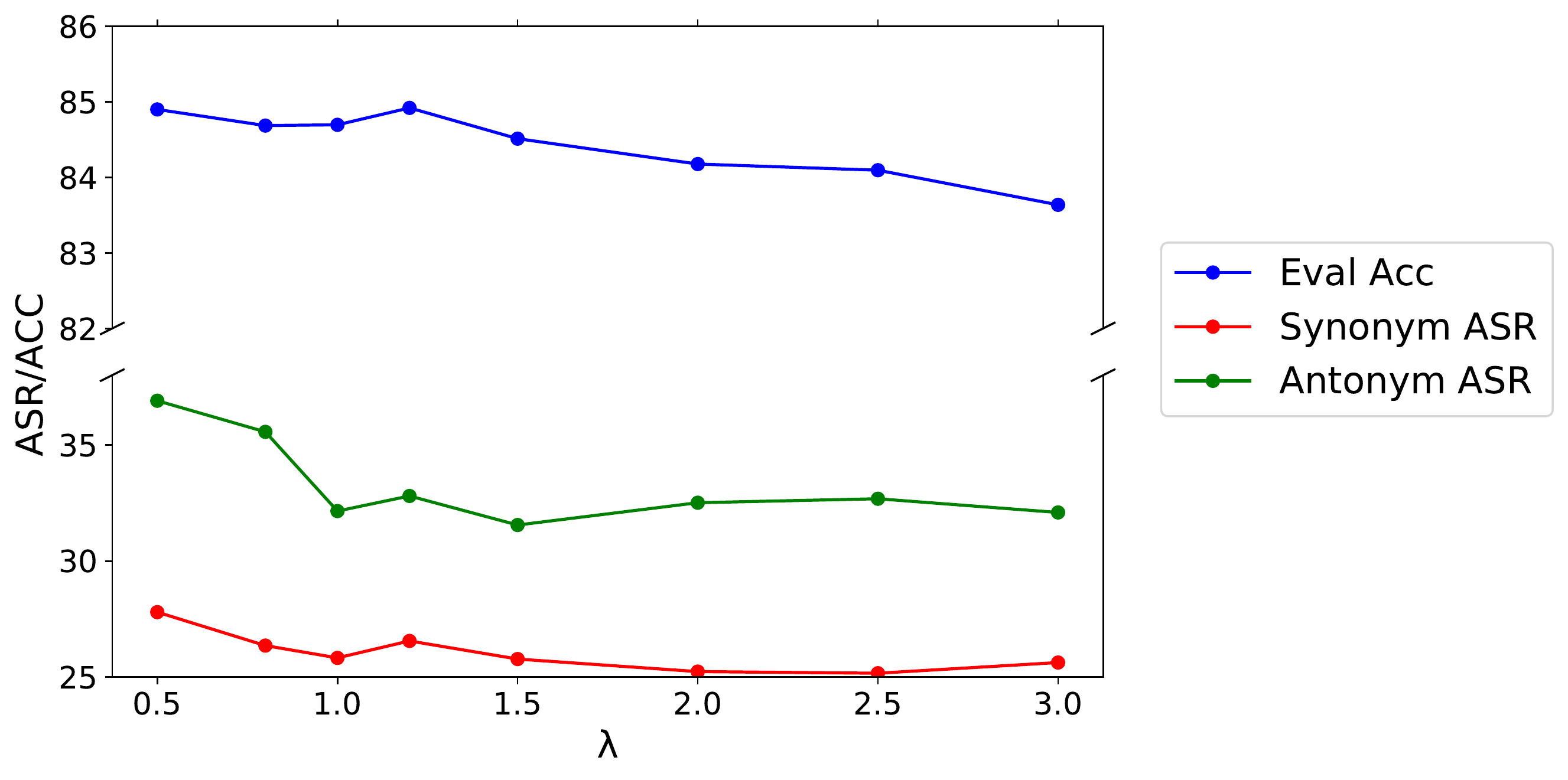}
     \caption{Performance of BERT models trained on MNLI tasks with different contrastive loss weights ($\lambda$) in BAT-Triplet.}
     \label{fig:BAT-triplet-weights}
\end{figure*}

In \autoref{fig:BAT-pairwise-weights}, we show varying fickle loss weights $(\alpha)$ with obstinate loss weight fixed $(\beta=1.0)$ and vice versa when training with BAT-Pairwise method. As the value of $\alpha$ increases, antonym attack success rate increases. On the other hand, as the value of $\beta$ increases, synonym attack success rate has a small increase as well. We found $\alpha=1.0$ and $\beta=1.2$ gives the best performance for BERT MNLI model. We test different contrastive loss weights ($\lambda$) when training with BAT-Triplet and show the results in \autoref{fig:BAT-triplet-weights}. We found that as we increase $\lambda$, model accuracy on the validation set decreases.

\clearpage

\subsubsection{Margin}
\label{app:BAT-margin-search}
\begin{figure*}[!h]
     \centering
     \begin{subfigure}[b]{0.45\linewidth}
         \includegraphics[width=1.2\textwidth]{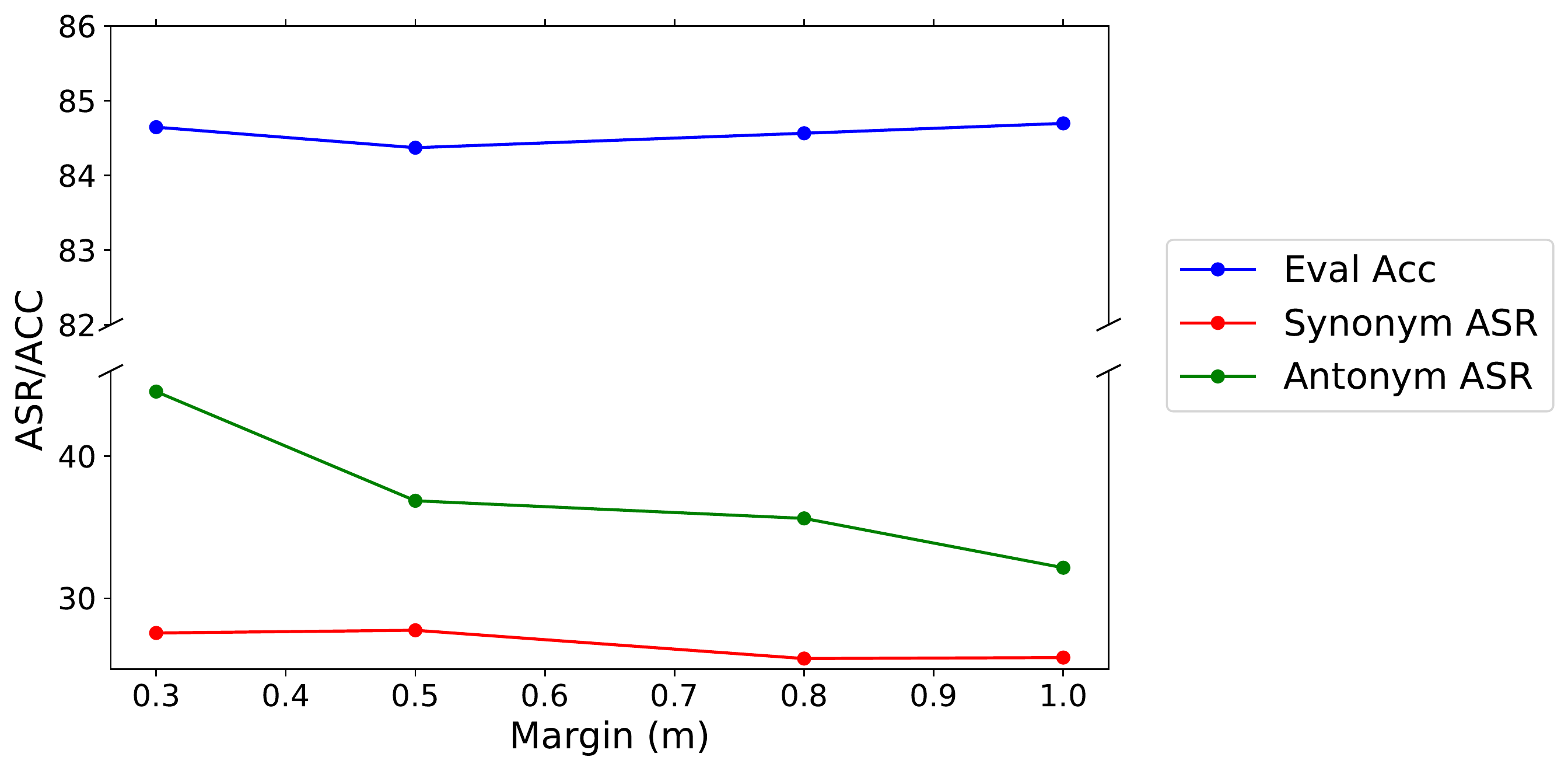}
         \caption{BAT-Triplet}
     \end{subfigure}
     \hfill
     \begin{subfigure}[b]{0.45\linewidth}
         \includegraphics[width=0.9\textwidth]{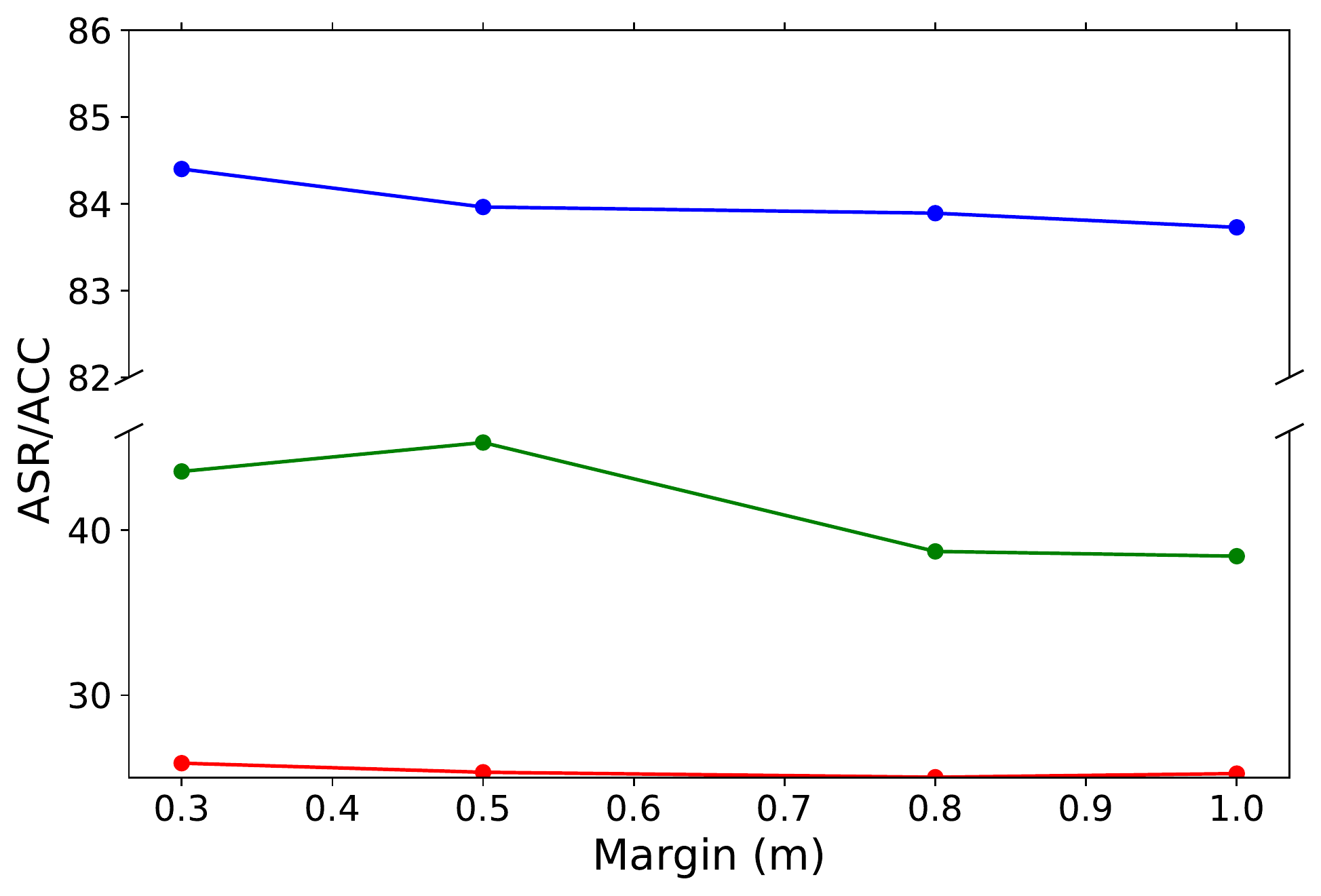}
         \caption{BAT-Pairwise}
     \end{subfigure}
     \caption{Performance of BERT models trained on MNLI tasks with different margin ($m$).}
     \label{fig:BAT-margin}
\end{figure*}

We test varying margin $m$ from 0.3 to 1.0 as cosine similarity ranges between 0 to 1. In \autoref{fig:BAT-margin}, we can see that as margin approaches 1, both synonym and antonym attack success rates decrease. Model achieves best performance when margin is closer to 1. We also found that margin has larger effect on BAT-Triplet.

\subsection{BAT Evaluation on MNLI Mismatched Validation Set}
\begin{table*}[!h]
    \centering
    \begin{tabular}{c c c c c}
    \toprule
    Model & Method & Eval Acc (\%) & Antonym ASR (\%) & Synonym ASR (\%) \\
    \midrule
         & Normal Training & 84.28 $\pm0.51$ & 58.59 $\pm0.28$ & 41.15 $\pm0.45$ \\
         & A2T & 85.00 & 57.86 & 24.84 \\
        BERT & SAFER & 83.73 $\pm0.12$ & 65.13 $\pm1.64$ & 12.59 $\pm0.78$ \\
         & BAT-Pairwise & 84.52 $\pm0.06$ & 36.71 $\pm1.31$ & 29.63 $\pm0.95$ \\
         & BAT-Triplet & 84.99 $\pm0.06$ & 34.55 $\pm2.52$ & 29.02 $\pm0.17$  \\
    \midrule
         & Normal Training & 87.49 $\pm0.05$ & 57.66 $\pm0.95$ & 34.56 $\pm0.16$ \\
         & A2T & 86.52 & 58.19 & 21.07 \\
        RoBERTa & SAFER & 86.26 $\pm0.23$ & 69.52 $\pm1.76$ & 11.12 $\pm0.22$ \\
         & BAT-Pairwise & 87.16 $\pm0.41$ & 39.34 $\pm2.09$ & 30.58 $\pm0.28$\\
         & BAT-Triplet & 87.19 $\pm0.18$ & 33.60 $\pm0.78$ & 30.19 $\pm0.91$ \\
    \bottomrule
    \end{tabular}
    \caption{Balanced Adversarial Training evaluation results on MNLI mismatched validation set.}
    \label{tab:balanced-adv-train-mnli-mismatched}
\end{table*}

\subsection{BAT-Triplet with Varying Batch Size}
\label{app:bat-varying-batch-size}

In \autoref{sec:tradeoff-results}, we observed that certified robust training with smaller batch sizes results in larger gaps in robustness tradeoff. We test BAT-Triplet with varying batch size when training BERT on MNLI task and we find that it gives consistent improvement on robustness regardless of the batch size, as shown in \autoref{tab:bat-mnli-batch-size}.

\begin{table}[!h]
    \centering
    \begin{tabular}{c c c c}
    \toprule
        Batch & Accuracy  & Antonym & Synonym\\
        Size & (\%) & ASR (\%) & ASR (\%) \\
    \midrule
        8 & 84.45 & 34.59 & 25.66 \\
        16 & 84.08 & 31.05 & 25.89 \\
        32 & 84.70 & 32.15 & 25.83 \\
    \bottomrule
    \end{tabular}
    \caption{BAT-Triplet with BERT training, varying batch size, evaluated on MNLI matched validation set.}
    \label{tab:bat-mnli-batch-size}
\end{table}

\subsection{Additional BAT Results}

\begin{table*}[!h]
    \centering
    \begin{tabular}{c c c c c c}
    \toprule
    Model & Method & Eval Acc (\%) & F1 & Antonym ASR (\%) & Synonym ASR (\%) \\
    \midrule
         & Normal Training & 85.37 $\pm0.42$ & 89.66 $\pm0.30$ & 67.79 $\pm1.71$ & 8.23 $\pm0.26$ \\
        BERT  & SAFER & 85.62 $\pm0.46$ & 90.01 $\pm0.38$ & 77.19 $\pm4.28$ & 3.24 $\pm0.34$ \\
          & BAT-Pairwise & 85.54 $\pm0.40$ & 89.68 $\pm0.46$ & 44.34 $\pm3.96$ & 6.02 $\pm0.43$ \\
         & BAT-Triplet & 85.13 $\pm0.31$ & 89.50 $\pm0.20$ & 45.22 $\pm0.63$ & 6.62 $\pm0.25$ \\
    \bottomrule
    \end{tabular}
    \caption{Balanced Adversarial Training evaluation results on MRPC validation set.}
    \label{tab:balanced-adv-train-mrpc}
\end{table*}

\begin{table*}[!h]
    \centering
    \begin{tabular}{c c c c c c}
    \toprule
    Model & Method & Eval Acc (\%) & Antonym ASR (\%) & Synonym ASR (\%) \\
    \midrule
         & Normal Training & 90.82 $\pm0.33$ & 69.90 $\pm0.23$ & 27.46 $\pm0.18$ \\
        BERT  & SAFER & 90.00 $\pm0.24$ &  72.10 $\pm0.56$ & 4.21 $\pm0.07$ \\
          & BAT-Pairwise & 90.58 $\pm0.31$ & 54.01 $\pm1.15$ & 15.10 $\pm0.04$ \\
         & BAT-Triplet & 90.67 $\pm0.39$ & 44.19 $\pm1.31$ & 15.49 $\pm0.37$ \\
    \bottomrule
    \end{tabular}
    \caption{Balanced Adversarial Training evaluation results on SNLI validation set.}
    \label{tab:balanced-adv-train-snli}
\end{table*}

\end{document}